\PassOptionsToPackage{table}{xcolor}
\documentclass[sigconf,screen,nonacm]{acmart}
\AtBeginDocument{%
  }

\usepackage{float}
\usepackage{enumitem}
\usepackage{multirow}
\usepackage{algorithm}
\usepackage{algpseudocode}
\usepackage{tabularx}
\usepackage{listings}
\usepackage{makecell}
\usepackage{booktabs}

\usepackage{afterpage}

\usepackage{amsmath,amsfonts,bm}

\def\eqref#1{equation~\ref{#1}}
\def\1{\bm{1}}

\DeclareMathAlphabet{\mathsfit}{\encodingdefault}{\sfdefault}{m}{sl}
\SetMathAlphabet{\mathsfit}{bold}{\encodingdefault}{\sfdefault}{bx}{n}

\newcommand{\modelname}{Vis-Lens}
\newcommand{\etal}{\textit{et al.}}
\newcommand{\repo}{\href{https://github.com/NeuroAIHub/Vis-Lens}{\textcolor{black}{\underline{https://github.com/NeuroAIHub/Vis-Lens}}}}

\title{Structured Visual Landscape: Generating Preferred Representations in Biological and Artificial Neural Networks}

\author{Yusong Zhao}
\email{yusongz@illinois.edu}
\affiliation{%
  \institution{University of Illinois Urbana-Champaign}
  \city{Urbana}
  \state{Illinois}
  \country{United States}
}

\author{Edgar Y. Walker}
\email{eywalker@uw.edu}
\affiliation{%
  \institution{University of Washington}
  \city{Seattle}
  \state{Washington}
  \country{United States}
}

\author{Hao Wang}
\email{hwangml@illinois.edu}
\affiliation{%
  \institution{University of Illinois Urbana-Champaign}
  \city{Urbana}
  \state{Illinois}
  \country{United States}
}

\author{Mi Lu}
\email{lmi7@gatech.edu}
\affiliation{%
  \institution{Georgia Institute of Technology}
  \city{Atlanta}
  \state{Georgia}
  \country{United States}
}
\renewcommand{\shortauthors}{Zhao et al.}

\begin{document}

\begin{abstract}
Understanding how neurons respond to visual stimulus is an important question in both deep learning and neuroscience. It has significant implications for enhancing the interpretability of black-box artificial neural networks and for understanding the visual representation in the human brain. We propose a structured visual representations landscape and design an activation-based prior that allows us to effectively regularize the landscape with activations from brain regions or neural network units. Our model \textbf{Vis-Lens} integrates a variational autoencoder and a diffusion model into an image generation framework.
It enables the generation of natural, realistic preferred images by directly modifying the activation-regularized latent codes without per-image iterative optimization. We demonstrate the effectiveness of our framework on both artificial neural networks and human visual cortex using functional magnetic resonance imaging (fMRI) data. Our framework improves activation modulation and image realism compared with a recent state-of-the-art method for generating preferred visual representations of these neural targets. Code is available at \repo.
\end{abstract}


\maketitle
\section{Introduction}
\label{sec:intro}

The exploration of visual representations in both biological and artificial neural networks has advanced our understanding of how complex visual stimuli are processed and represented in neural networks~\citep{marr2010vision, hubel1962receptive, bashivan2019neural}.
Previous studies have demonstrated the capacity of neurons in the medial temporal lobe (MTL) of the human brain to form invariant representations of complex stimuli~\citep{quiroga2005invariant}, such as faces, landmarks, and objects, regardless of visual variations. These findings illustrate how single neurons can encode high-level, abstract percepts with remarkable specificity, leading to the hypothesis of sparse and invariant coding mechanisms in the brain~\citep{olshausen2004sparse}.

\begin{figure*}[t]
  \centering
   \includegraphics[width=0.9\linewidth]{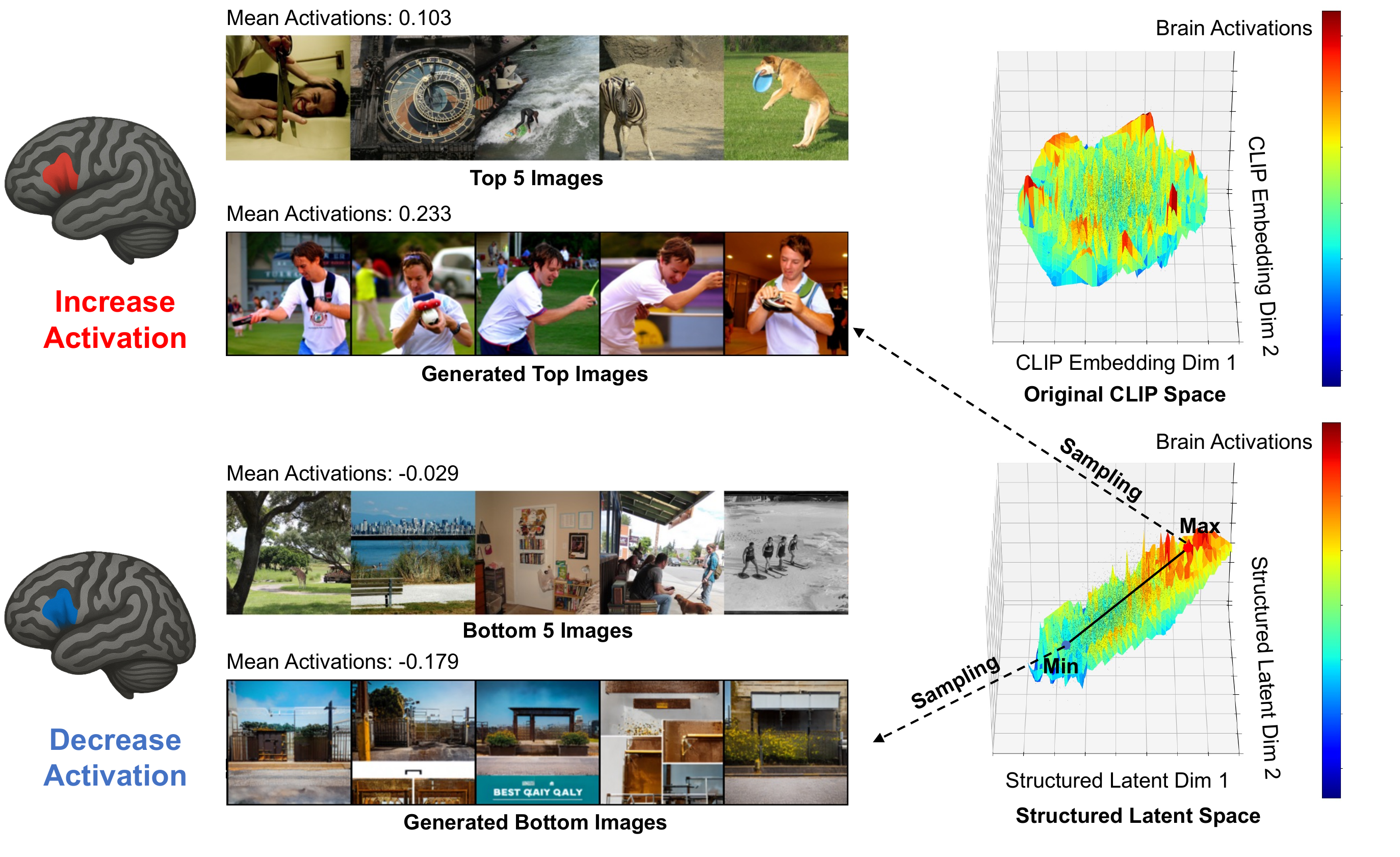}
   \caption{\textbf{Original CLIP Space (Top):} Original CLIP embedding space without any activation constraint: image embeddings that elicit high and low neural responses are dispersed and not clustered throughout the space without shared visual structure. \textbf{Structured Landscape (Bottom):} By encoding CLIP embeddings into the regularized latent space constrained with neuronal activations (after z-score normalization) as the \textit{prior}, our method \modelname{} could provide a more structured and representative landscape to better identify the representations and preferences of brain regions than sorting top-k images in a dataset. The results are based on brain activations from the NSD dataset~\citep{Allen2022}. In the landscape plots, the color bar and z axis indicate the activation of the target brain region given different images as inputs, while the x and y axes represent two latent dimensions, respectively.}
   \label{fig:landscape}
\end{figure*}

While methods exploring visual representation in some brain regions work fine by learning directly from embeddings of popular image encoders like CLIP~\citep{garciacerdas2025brainactiv, luo2023braindiffusionvisualexploration}, understanding the feature representations for many brain regions that are not well-studied remains challenging. For example, some higher-order brain regions can demonstrate mixed selectivity~\citep{Rigotti2013The}, where they can respond to images with small overlap in shared features. As presented in Figure~\ref{fig:landscape}, the top images that maximally activate the brain region are distributed across the embedding space of the CLIP encoder~\citep{radford2021clip}.

Additionally, brain activity measured is often noisy, low-resolution, and partially observed, which introduces difficulties for analysis and modeling. Unlike previous works that aim to reconstruct entire images from brain activities~\citep{naselaris2011encoding, nishimoto2011reconstructing, shen2019deep, horikawa2017generic}, our work focuses on understanding the feature representation of brain regions by generating new preferred images that maximize or minimize the activity of those regions. These synthetic visual stimuli can be deployed in follow-up neuroscience experiments to test new hypotheses, refine ROI functional maps, and accelerate discovery by guiding stimulus design.

To achieve this goal, we propose \modelname{}, an effective approach to interpret the feature represented by brain regions. Given the noisy nature of brain recordings, instead of using brain activities directly as input to the decoding model, our method refines the landscape of visual representation by employing brain activation to regularize the prior, serving as a soft constraint. We further integrate the structured landscape with a variational auto-encoder (VAE) and diffusion generative models to generate realistic preferred images that can effectively modulate target brain activations.

Our method can facilitate the exploration of visual representations of brain regions while maintaining interpretability. It is validated with fMRI data from the Natural Scenes Dataset (NSD)~\citep{Allen2022}, analyzing responses in the human visual cortex to complex natural stimuli. Meanwhile, due to the scarcity of image response data from human brains, our proposed framework is also extended to artificial neural networks (ANNs), allowing controlled testing of visual feature representations and interpretability in artificial systems. By integrating insights from both biological and artificial neural networks, we aim to bridge the understanding of visual representation strategies across natural and engineered systems. Our contributions can be outlined as follows:

\begin{itemize}
\item
We introduce an activation-regularized prior for VAE~\citep{kingma2013auto} that organizes its latent space into two distinct clusters, corresponding to codes that increase or decrease a target brain region's activation. This design enables easy plug-and-play integration with conventional generative models.

\item
Our method synthesizes visual representations from specified brain regions, showing better performance to regularize brain activations than a recent state-of-the-art baseline, especially for decreasing activations. It could also generate preferred visual inputs with better perceptual realism. Moreover, it provides a generalized framework which shows robustness on cross-subject biological data, and can also be transferred to visualizing features in artificial neural networks.

\item
By generating controllable, high-quality visual stimuli that can modulate specific regions, our framework demonstrates new potential in vision-based brain computer interface applications and can be used to generate new hypotheses for future neuroscience studies.
\end{itemize}

\section{Related Works}
\label{sec:Related works}
\noindent\textbf{Image generative models.}
Advances in generative models have significantly transformed image synthesis. Foundational approaches such as GANs~\citep{goodfellow2014generative} led to extensions like Wasserstein GANs~\citep{arjovsky2017wasserstein} and Conditional GANs~\citep{mirza2014conditional}, with fidelity improvements using spectral normalization~\citep{miyato2018spectral}, while VAEs~\citep{kingma2013auto} introduced probabilistic encoding, later combined with latent space regularization~\citep{ho2022cascaded, van2016conditional}. Further, Cascaded generation~\citep{ho2022cascaded} and autoregressive modeling~\citep{chen2018pixelsnail} enhanced synthesis quality. More superior generation models like Diffusion models~\citep{ho2020denoising, rombach2022highresolutionimagesynthesislatent} advanced controllable, high-fidelity synthesis, integrating structured perturbations via SDEdit~\citep{meng2022sdeditguidedimagesynthesis} and text-to-image alignment through CLIP~\citep{radford2021clip}. IP-Adapter~\citep{ye2023ipadaptertextcompatibleimage} refined latent spaces for semantic consistency, while GLIDE~\citep{nichol2022glidephotorealisticimagegeneration} and Stable Diffusion~\citep{rombach2022highresolutionimagesynthesislatent} further demonstrated photorealistic, text-guided image generation.
Our work is developed based on these conventional generative model frameworks, it adopts VAEs to build up our proposed structured landscape and a Stable Diffusion model generation pipeline to synthesize preferred images conditioned on CLIP embeddings derived from the structured landscape.

\noindent\textbf{Feature visualization.} Visualization techniques provide deep insights into neural network interpretability and functionality, spanning both biological and artificial systems. Early work by Erhan \etal~\citep{erhan2009visualizing} focused on minimal regularization approaches, while adversarial examples exposed vulnerabilities in neural networks~\citep{szegedy2013intriguing}. Mahendran and Vedaldi~\citep{mahendran2015understanding} and Nguyen \etal~\citep{nguyen2015deep} further expanded on interpretability with total variation regularization, counterexamples, and image blurring. Meanwhile, techniques such as DeepDream by Mordvintsev \etal~\citep{mordvintsev2015inceptionism, mordvintsev2016deepdream}, employing jitter, multi-scale visualization, and gradient normalization~\citep{oygard2015blurring, tyka2016regularization}, enhanced clarity in feature visualization. For prior related work, Nguyen \etal~\citep{nguyen2016synthesizing} demonstrated generative adversarial synthesis of preferred inputs, emphasizing controlled image generation through learned priors. Our work extends this line by generating images with preferred features based on pre-defined priors in landscape shaped by activations.

\noindent\textbf{Brain decoding and most exciting images.}
Generative models leveraging brain activity as direct input have substantially advanced neural-to-visual synthesis. Direct neural-to-image synthesis methods convert fMRI activity into high-fidelity pictures with diffusion or GAN backbones, like BrainDiffusion and NeuroDM \citep{luo2023braindiffusionvisualexploration, Zeng2024NeuroDM}, Brain2GAN \citep{Dado2024Brain2GAN}, latent-diffusion variants \citep{Ferrante2024NeuroInspired, ozcelik2023naturalscenereconstructionfmri}, and energy-guided approaches \citep{ExcitingImages2023}, while activation-optimized methods like Inception Loops~\citep{Inception2019} and NeuroGen~\citep{GU2022118812} reveal neuronal tuning by iterating stimuli based on neural responses. Recent brain decoding studies have continued to improve visual reconstruction quality through stronger priors, better alignment spaces, and cross-subject or data-efficient formulations, including generative decoding with latent diffusion~\citep{Takagi2023, ozcelik2023naturalscenereconstructionfmri}, diffusion-prior-based reconstruction~\citep{scotti2023reconstructingmindseyefmritoimage, scotti2024mindeye2sharedsubjectmodelsenable}, probabilistic visual-to-fMRI synthesis~\citep{mai2025synbrainenhancingvisualtofmrisynthesis}, and cross-subject decoding frameworks~\citep{wang2024mindbridgecrosssubjectbraindecoding}. Recent multi-modal approaches~\citep{VisualDecoding2021, Ding2023Bipartite, vangerven2021neural, qiu2025mindllmsubjectagnosticversatilemodel, benara2024craftinginterpretableembeddingsasking, huo2026pictorialcortexzeroshotcrosssubject} integrate generative models for improved synthesis precision, while Controllable Mind Visual Diffusion~\citep{zeng2023controllablemindvisualdiffusion}, Seeing Beyond the Brain~\citep{chen2023seeingbrainconditionaldiffusion}, and Reconstructing the Mind's Eye~\citep{scotti2023reconstructingmindseyefmritoimage} leverage structured diffusion priors for enhanced visual reconstructions. Beyond reconstruction, recent methods have also explored richer alignment spaces between brain signals and foundation-model representations, such as CLIP-based decoding~\citep{liu2023} and self-supervised visual representations including DINO-style encoders~\citep{Benchetrit2024}. Though dominated by fMRI data, ~\citep{song2024decodingnaturalimageseeg, li2024visualdecodingreconstructioneeg, EGG2024, EEG2024Ma, li2025brainflorauncoveringbrainconcept, Abramov_2025} explored on EEG data, obtained promising results. 

To further modulate activations, computational models~\citep{Papale2024.10.09.617382, Murty2021Computational} and frameworks targeting single neuron contributions~\citep{bau2017network, olah2017feature} provide interpretable mappings of cortical activations to specific visual features. Latent space optimization strategies~\citep{NeuralDecoding2023, xia2024dreamvisualdecodingreversing} extend foundational decoding frameworks~\citep{naselaris2011encoding, nishimoto2011reconstructing, shen2019deep, horikawa2017generic} to achieve targeted neural activation modulation.  BrainDiVE and BrainSCUBA\citep{luo2023braindiffusionvisualexploration, luo2024brainscubafinegrainednaturallanguage} uses a CLIP-based encoder and diffusion models to improve stimulus quality and semantic specificity. BrainACTIV~\citep{garciacerdas2025brainactiv}, as the recent state-of-the-art and the first work on using diffusion models to generate images to regulate brain activations, extends these paradigms by conditioning synthesis with linearly fitted CLIP embeddings and brain activation patterns. Related efforts in controllable activation synthesis have also shown that generative models can be used not only to maximize neural responses but also to suppress or differentially modulate them, both in biological systems and in artificial networks~\citep{bashivan2019neural, Shinkle2025, Nguyen2017}. Although modeling activations via CLIP embeddings does work in previous work, our method tries to take the noisy and mix-selective nature of brain activity data into consideration, building a novel structured landscape in VAE through CLIP embeddings, permitting more controllable and effective activation manipulation, and avoiding falling into local minimum in the optimization process.

\begin{figure*}[t]
  \centering
  \includegraphics[width=\linewidth]
  {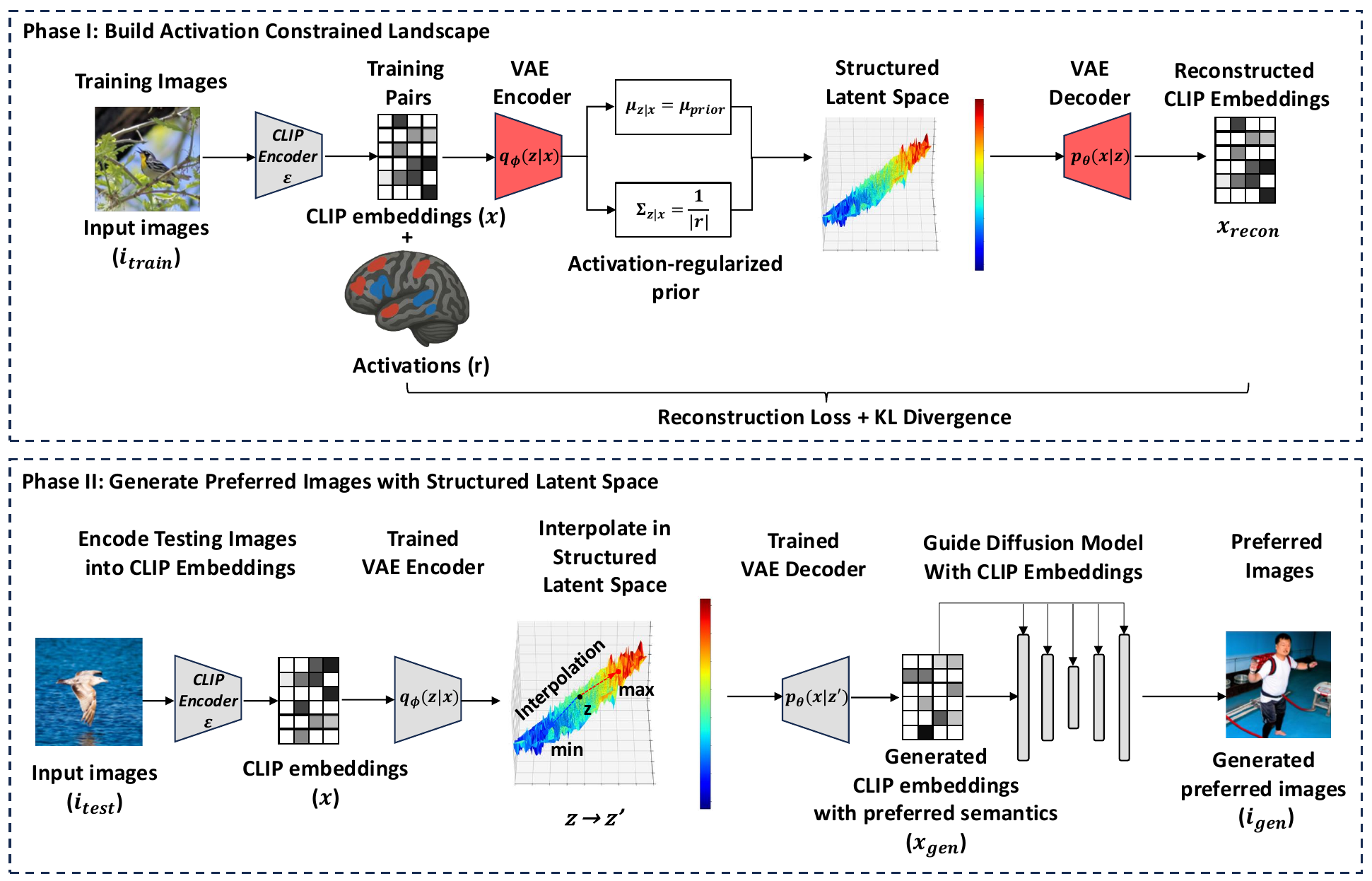}
  \vspace{-10pt}
   \caption{\textbf{Overview of \modelname{}}. The illustration depicts a two-stage approach integrating \textbf{a structured landscape and variational autoencoder (VAE) generation} with CLIP encoder and diffusion based generative model. \textbf{Phase I:} The input image, $i_{\text{train}}$, is encoded by a pretrained CLIP encoder $\mathcal E$, CLIP embeddings $x$ are encoded into a latent representation $z$ by the VAE encoder. A regularization step using a \textbf{activation-based prior} $\mathcal{N}(\mu_\text{prior}, \frac{1}{\mid r\mid})$ modifies the latent space for better alignment with activations, producing a regularized latent $z$. The VAE is also trained to reconstruct the input CLIP embeddings $x$ with $x_{recon}.
 $~\textbf{Phase II:} The test images $i_{\text{test}}$, is encoded by same CLIP encoder $\mathcal E$. Then, with fully trained VAE and the structured visual latent representation space, we can directly \textbf{modify latents} constrained by a specific feature represented by the selected regions, obtaining a new latent $z'$. This transformed latent is decoded back into an intermediate generated CLIP embeddings $x_{gen}$, which is used to guild the image generation through a diffusion pipeline, yielding the final output  $i_{\text{gen}}$ as preferred images to modulate the activation $r$.}
   \label{Method_Overview}
\end{figure*}

\section{Methodology}
\label{sec:methodology}

\subsection{Problem Setting and Notation}
\label{subsubsec:Problem Setting and Notation}
Let $\mathcal I$ be the space of natural images, $\mathcal X$ be the space of CLIP embeddings, and $\mathcal Z$ be the latent space of a variational auto-encoder (VAE).
For an input image $i\in\mathcal I$, the pretrained CLIP encoder $\mathcal E$ maps $i$ to a CLIP embedding $x = \mathcal E(i) \in \mathcal X$.
Subsequently, the VAE encoder $q_{\phi}(z\!\mid\!x)$ produces a latent
$z\in\mathcal Z$ with mean $\mu$ and diagonal covariance $\Sigma=\operatorname{diag}(\sigma^2)$;
the VAE decoder $p_{\theta}(x\!\mid\!z)$ reconstructs an embedding $x_\text{rec}$.
Throughout, $r$ denotes a scalar activation value of a chosen target, i.e., the mean response of a selected ROI in human brain data after z-score normalization or the response of a selected unit in an artificial neural network.
$\lambda,\,\omega_{\text{rec}},\,\omega_{\text{KLD}}$ are scalar hyper-parameters.

\subsection{Constructing a Structured VAE Landscape with Activation-Constrained Priors}
\label{subsubsec:Build Structured Landscape with Neuronal Activation Regularization}

To generate explainable and controllable preferred visual representations for a specific brain region or neural unit, we incorporate the target activation $r$ into the VAE prior and use it to organize the latent space.
The key intuition is that neural signals are noisy and could have an asymmetric distribution. To better represent those properties, we design a prior for the neural responses that they have a bimodal distribution, with significant responses (most positive or negative) are expected to be centralized in each distribution. The greater the absolute value of activation is, the closer it will be to each distribution center, and responses close to zero are relatively loosely constrained.
This design yields a structured latent landscape that supports direct latent manipulation at test time.

The KL divergence (KLD) term in the VAE measures the divergence between the encoder's approximate posterior, $q_\phi(\mathbf{z}|\mathbf{x})$, and the prior, $p_\theta(\mathbf{z})$. We begin with its standard definition:
\begin{equation}
\label{eq:kld_definition}
\begin{split}
D_{\mathrm{KL}}(q_\phi(\mathbf{z}|\mathbf{x}) \parallel p_\theta(\mathbf{z}))
&= \int q_\phi(\mathbf{z}|\mathbf{x}) \log \frac{q_\phi(\mathbf{z}|\mathbf{x})}{p_\theta(\mathbf{z})}
\, d\mathbf{z} \\
&= \mathbb{E}_{\mathbf{z}\sim q_\phi(\mathbf{z}|\mathbf{x})}
\left[ \log q_\phi(\mathbf{z}|\mathbf{x}) - \log p_\theta(\mathbf{z}) \right].
\end{split}
\end{equation}

Both posterior and prior are defined as multivariate Gaussian
distributions with diagonal covariance matrices. In our framework, the
approximate posterior is
$q_\phi(\mathbf{z}\mid\mathbf{x})
= \mathcal{N}(\mathbf{z}; \boldsymbol{\mu}, \operatorname{diag}(\boldsymbol{\sigma}^2))$.
For the prior, we define an \textbf{activation-constrained prior:}
\[
p_\theta(\mathbf{z}) = \mathcal{N}(\mathbf{z}; \boldsymbol{\mu}^{\text{prior}}, \operatorname{diag}((\boldsymbol{\sigma}^{\text{prior}})^2)),
\]
which depends on the target activation value $r$:
\begin{equation}
\label{eq:priors}
\sigma_i^{\text{prior}} = \left| \frac{\lambda}{r_i + \epsilon} \right|,
\qquad
\mu_i^{\text{prior}} =
\begin{cases}
\mu_{\text{pos}}, & \text{if } r > 0, \\
\mu_{\text{neg}}, & \text{otherwise}.
\end{cases}
\end{equation}

Here, $\lambda$ is a scaling hyperparameter, $\epsilon$ is a small constant to prevent division by zero, and $\mu_{\text{pos}}$, $\mu_{\text{neg}}$ are predefined means for positive and negative activations, respectively.
The role of $\mu_{\text{pos}}$ and $\mu_{\text{neg}}$ is to provide two separated latent centers in VAE latent space for maximize activation increase and decrease.
Meanwhile, variance term $\sigma^{\text{prior}}_i$, which defines different priors for each data point $i$, decrease as $|r_i|$ increases. By imposing such continuous loose to tight constraints for each data point, our method can map images to a structured space. Theoretically, the inputs will be centralized when $r_i$ approaches infinity, while in practice, we can sample input data points with maximize and minimize values at two pre-defined centers, and they are expected to increase or decrease the activations of targeted neurons.

Given the definition of the approximate posterior and activation-constrained prior, the expectation in Eq.~\ref{eq:kld_definition} can be computed analytically, yielding the closed-form KLD between two Gaussian distributions. Assuming that the VAE latent space has $d_z$ dimensions, the final activation-aware KLD is
\begin{equation}
\label{eq:kld_modified}
\text{KLD}_{\text{activation}} = \sum_{i=1}^{d_z} \left[ \log \frac{\sigma_i^{\text{prior}}}{\sigma_i} - \frac{1}{2} + \frac{\sigma_i^2 + (\mu_i - \mu_i^{\text{prior}})^2}{2 (\sigma_i^{\text{prior}})^2} \right].
\end{equation}

Here, $\mu_i$ and $\sigma_i$ are the mean and standard deviation of the approximate posterior for the $i$-th latent dimension, respectively; $\mu_i^{\text{prior}}$ and $\sigma_i^{\text{prior}}$ are the corresponding mean and standard deviation of the prior.
This modified KLD does more than regularize the latent space toward a standard Gaussian: it explicitly organizes latent codes according to activations. As a result, the learned latent space becomes more suitable for controllable activation modulation than a single linear direction in the original embedding space.

With this activation-constrained prior, we can build a structured VAE latent space, where we can easily perform in-distribution sampling from a spherical space in a structured landscape.

\begin{figure*}[t]
  \centering
  \vspace{-5pt}
  \includegraphics[width=\linewidth]{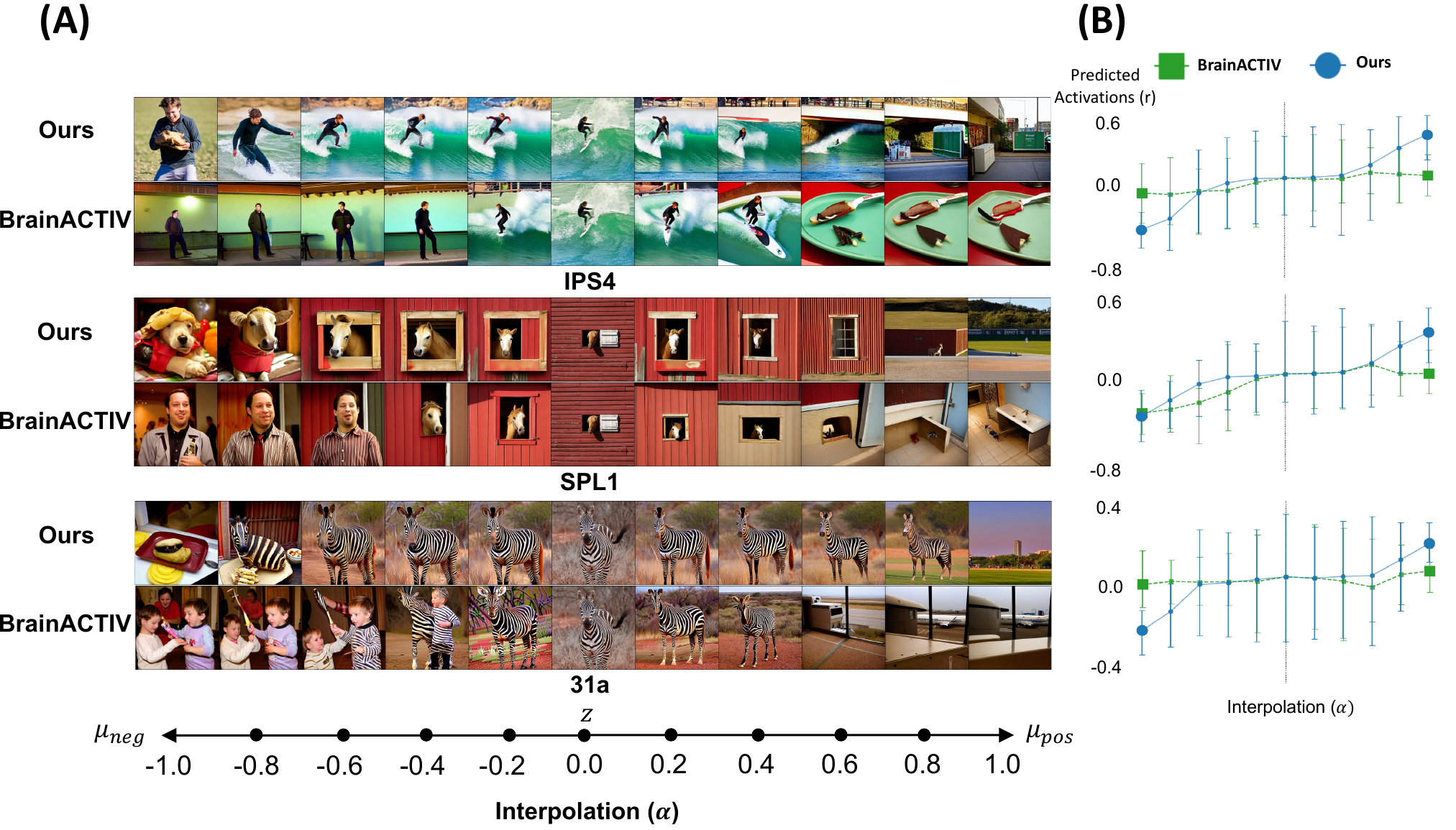}
  \vspace{-10pt}
   \caption{\textbf{Activation change results of our method and BrainACTIV \cite{garciacerdas2025brainactiv} (A)} Generate preferred representations of multiple ROIs with example images from COCO dataset by our method and baselines. The alpha value in interpolation indicates the strength and direction of preferred semantic in generation. \textbf{(B)} Comparison between predicted activations of ROIs made by DINO-ViT encoder. Means and ranges of activations are presented for different alpha settings. Our method outperforms the baseline in terms of maximizing or minimizing the ROI activations.}
   \Description{Activation change results of our method and baseline, showing generated images and predicted activations for different ROIs.}
   \label{fig:NSD_Result}
\end{figure*}

\subsection{Generate Preferred Visual Representations with Modified VAE}
\label{sec:training_and_generation}

To produce preferred images, our method consists of two phases: a training phase that constructs the structured latent space, and a generation phase that synthesizes images from it. The VAE is the only trainable component in the pipeline, while the pretrained CLIP encoder ($\mathcal{E}$) and the image-to-image diffusion model remain frozen.

\noindent\textbf{Training.}
During training, only the VAE encoder $q_{\phi}(z\!\mid\!x)$ and decoder $p_{\theta}(x\!\mid\!z)$, which map the CLIP embedding to a latent representation $z$ and reconstruct it back to $x_{\mathrm{recon}}$, are optimized by
\begin{equation}
    \label{eq:vae_loss}
    \mathcal{L}_{\text{VAE}} = \omega_{\text{recon}} \cdot \mathcal{L}_{\text{recon}} + \omega_{\text{KLD}} \cdot \text{KLD}_{\text{activation}},
\end{equation}
where $\mathcal{L}_{\text{recon}} = \lVert x - x_{\text{recon}} \rVert_2^2$ is the reconstruction loss between the original and decoded CLIP embeddings. The terms $\omega_{\text{recon}}$ and $\omega_{\text{KLD}}$ balance reconstruction fidelity against the activation-based regularization imposed by Eq.~\ref{eq:kld_modified}.
Therefore, the VAE is trained not only to reconstruct CLIP embeddings, but also to reshape the latent space such that neural activation information becomes geometrically accessible for later manipulation.

\noindent\textbf{Generation.}
To synthesize new images, we first encode an input image $i$ with the frozen CLIP encoder to obtain its CLIP embedding $x = \mathcal E(i)$. We then feed $x$ into the trained VAE encoder to obtain a latent code $z = q_{\phi}(x)$ in the structured latent landscape. Next, $z$ is moved toward the predefined positive or negative prior center, producing a modified latent code $z'$. The modified latent code is decoded into a new CLIP embedding $x' = p_{\theta}(x\,|\,z')$, which is then passed to the frozen diffusion model to synthesize the final preferred image $i'$.

\section{Experiments}

\begin{table*}[t!]
\centering
\caption{Comparison of activation change and realism between our method and the baseline on NSD datasets. Better values are in \textbf{bold face}. For activation changes, $\uparrow$ indicates that higher values are better, and $\downarrow$ indicates that lower values are better. Our method produces activation changes that are more consistently aligned with the intended direction.}
\label{tab:combined_brain_results}

\textbf{(a) NSD Results}
\resizebox{0.9\textwidth}{!}{%
\begin{tabular}{rlccccccc}
\toprule
& \textbf{ROIs} & \textbf{TE2p} & \textbf{IP0} & \textbf{IP1} & \textbf{OFA} & \textbf{FFA} & \textbf{EBA} & \textbf{OPA} \\
\midrule
\multirow{2}{*}{Activation Increase ($\uparrow$)} & Our Method & \textbf{0.242} & \textbf{0.498} & \textbf{0.153} & 1.052 & 0.891 & 0.963 & \textbf{0.699} \\
& BrainACTIV & 0.207 & 0.463 & 0.044 & \textbf{1.155} & \textbf{0.976} & \textbf{1.160} & \textbf{0.699} \\
\midrule
\multirow{2}{*}{Activation Decrease ($\downarrow$)} & Our Method & \textbf{-0.136} & \textbf{-0.699} & \textbf{-0.073} & \textbf{-0.660} & \textbf{-0.580} & \textbf{-0.810} & \textbf{-0.824} \\
& BrainACTIV & 0.149 & 0.186 & 0.092 & 0.213 & -0.484 & -0.669 & -0.172 \\
\midrule
\multirow{2}{*}{FID @ 2k ($\downarrow$)} & Our Method & \textbf{172.71} & \textbf{173.34} & \textbf{178.01} & \textbf{173.66} & \textbf{153.70} & \textbf{137.33} & \textbf{166.01} \\
& BrainACTIV & 225.77 & 213.17 & 200.14 & 216.15 & 256.44 & 188.59 & 225.37 \\
\bottomrule
\end{tabular}%
}

\end{table*}

\begin{table*}[h]
\centering
\caption{Cross-subject generalization results on the NSD dataset. The table compares predicted activation changes obtained from the original subject-specific predictor and a held-out-subject predictor across several visual ROIs.}
\label{tab:cross_subject}

\resizebox{0.9\textwidth}{!}{%
\begin{tabular}{lccccccc}
\toprule
\textbf{ROIs} & \textbf{OFA} & \textbf{FFA} & \textbf{EBA} & \textbf{VWFA} & \textbf{OPA} & \textbf{PPA} & \textbf{RSC} \\
\midrule
Activation Increase (original subj.) & 0.869 & 0.851 & 0.860 & 0.867 & 0.731 & 0.966 & 0.942 \\
Activation Increase (held-out subj.) & 0.861 & 0.975 & 0.790 & 0.648 & 0.629 & 1.061 & 0.940 \\
\midrule
Activation Decrease (original subj.) & -0.526 & -0.382 & -0.564 & -0.279 & -0.557 & -0.610 & -0.775 \\
Activation Decrease (held-out subj.) & -0.436 & -0.302 & -0.391 & -0.089 & -0.485 & -0.813 & -0.823 \\
\bottomrule
\end{tabular}%
}
\end{table*}

\label{sec:Experiments}
\subsection{Datasets}
\label{subsec:Datasets}
\textbf{Natural Scenes Dataset (NSD).} The Natural Scenes Dataset (NSD), introduced by Allen et al.~\cite{Allen2022}, provides high-resolution fMRI recordings from eight subjects viewing approximately 10,000 MS COCO images~\cite{lin2015microsoftcococommonobjects}, each repeated three times to enhance data consistency. Following standard fMRI encoding practices~\cite{naselaris2011encoding, VANGERVEN2017172}, beta values are z-scored within sessions and averaged across repetitions. ROI masks are utilized to extract activation patterns from specific brain regions, serving as essential constraints in our framework to align the VAE latent space with observed brain responses in ~\ref{subsubsec:Manipulate Activations in Human Brain}.

\noindent\textbf{ImageNet-mini.} ImageNet-mini is a subset of the full ImageNet dataset~\cite{5206848}, offering a diverse collection of images across various categories. This dataset serves as a benchmark for evaluating the performance of modulating artificial units from SIMCLR~\cite{chen2020simple} in ~\ref{subsubsec:Activate Unit in Artificial Neural Networks} and validating optimization performance under constrained image dimensions in ~\ref{subsubsec:Optimizing Unit Activations while Preserving Local Features}.

\subsection{Models}
\label{subsec:Models}
\textbf{Implementation of our framework.} Our VAE uses a linear encoder to map CLIP embeddings into an activation-constrained latent space and a linear decoder to reconstruct CLIP embeddings for controlled image synthesis. We use a pretrained CLIP-ViT-H-14 encoder~\citep{radford2021clip} to extract 1024-dimensional embeddings, and Stable Diffusion v1.5 with IP-Adapter~\citep{ye2023ipadaptertextcompatibleimage} as the image generator.

\noindent\textbf{Baseline method.}
For every ROI, we re-implement \emph{BrainACTIV}~\citep{garciacerdas2025brainactiv}, a ridge-regularized \emph{linear} model on \emph{$\ell_2$-normalized} CLIP embeddings, $[\; \tfrac{x}{\lVert x\rVert} \cdot\mathbf w \;+\; b \;]\;\Longrightarrow\;r$, yielding a single modulation vector $\mathbf w\!\in\!\mathbb R^{1024}$. This vector is interpreted as the direction of maximal and minimal activation in CLIP space: $\mathbf z_{\max} = \mathbf w / \lVert \mathbf w \rVert$ and $\mathbf z_{\min} = -\mathbf w / \lVert \mathbf w \rVert$. At inference time, a test CLIP embedding is shifted \emph{only} along $\mathbf z_{\max}$ or $\mathbf z_{\min}$ before being injected, via an unchanged adapter, into a frozen image-to-image diffusion model.
This baseline therefore assumes that activation increase and activation decrease can be represented by a single symmetric direction in CLIP space, whereas our method models them independently.

\begin{figure*}[h!]
  \centering
  \vspace{-5pt}
  \includegraphics[width=1\textwidth]{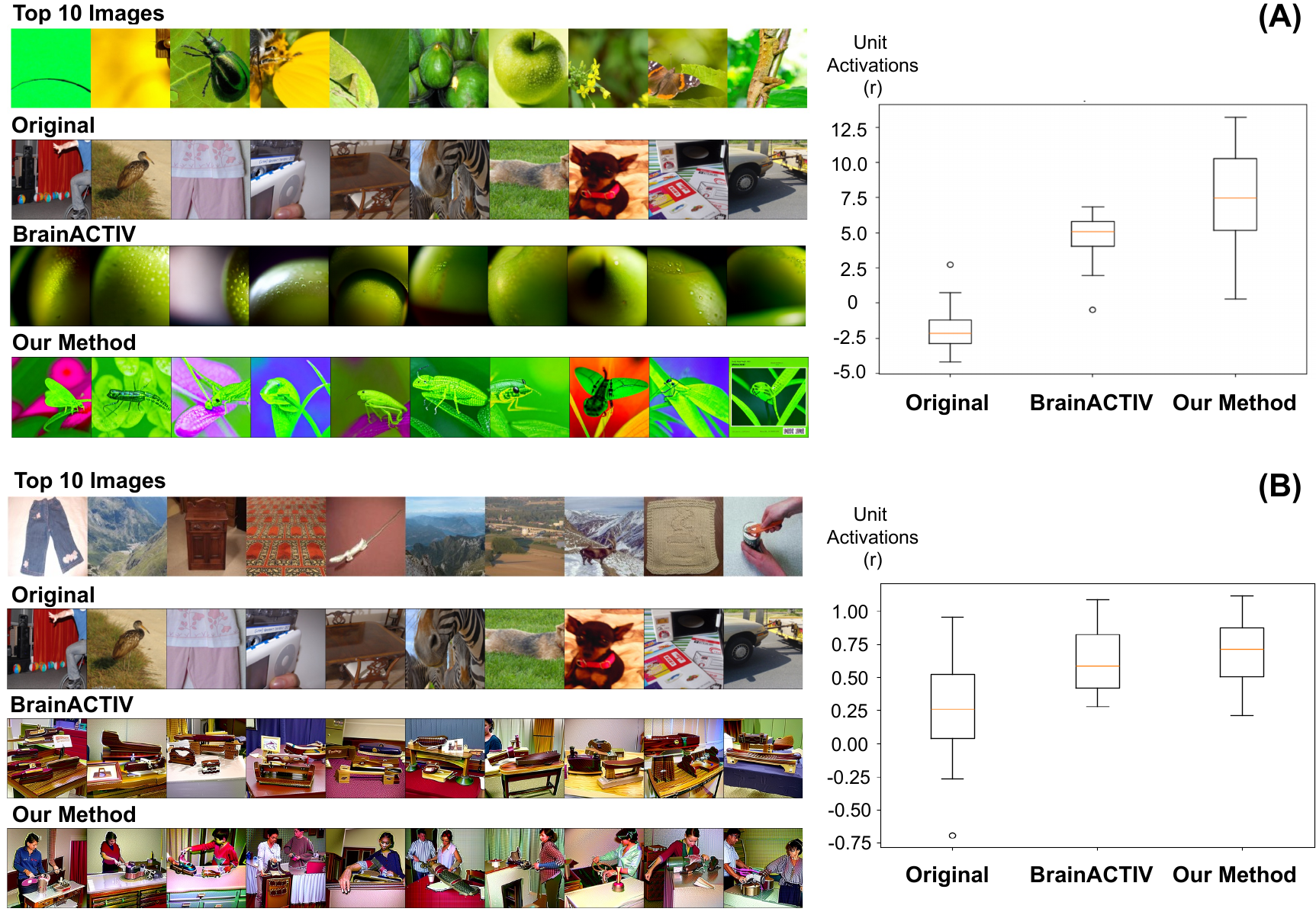}
  \vspace{-10pt}
   \caption{\textbf{Generate preferred images with increasing unit activations in ANNs. (A)} The top 10 images demonstrate a notable pattern of ``green plant \& insect'' represented by a unit in the middle layer of the SimCLR model. Our method generates images that better match this pattern and increase activation more than the BrainACTIV \cite{garciacerdas2025brainactiv}. \textbf{(B)} For a mid-layer unit in a more complex Vision Transformer (ViT), our method synthesizes images that can better activate the unit compared with the baseline, although the underlying concept appears more ambiguous.}
   \label{fig:ANN_activation_manipulation}
\end{figure*}

\noindent\textbf{In-silico brain simulator.}
In our experiments, brain activations are evaluated with a \emph{DINO-ViT} encoder~\citep{oquab2024dinov2learningrobustvisual}. DINO acts as a frozen feature extractor whose 12 layer outputs are fed to a single-layer ViT ensemble trained with ridge loss to predict neuronal responses. In line with the findings of \cite{garciacerdas2025brainactiv}, this encoder explains a large proportion of NSD voxel variance (typically $R^2\!>\!0.6$ in high-level ROIs), validating its use as a reliable neural readout. Because the encoder is not CLIP-based, it provides an independent estimate of neural activity.

\subsection{Results}
\label{subsec:Results}

\subsubsection{Generate Preferred Representations in Human Brain}
\label{subsubsec:Manipulate Activations in Human Brain}
\leavevmode\par
With the structured landscape, we can obtain a better representation of a target brain region by sampling around the corresponding prior in latent space and generating images from those latent values, as shown in Figure~\ref{fig:landscape}.

In our latent space, we interpolate each latent code $z$ toward a positive or negative prior,
$z'=\alpha\cdot \mu_{\text{prior}}+(1-\alpha)\cdot z$,
and pass it to a frozen SDEdit diffusion model with $\gamma=1.0$; thus, the output image depends solely on the edited embedding, allowing a clean comparison with the baseline.

\begin{figure*}[t]
  \centering
  \includegraphics[width=\textwidth]{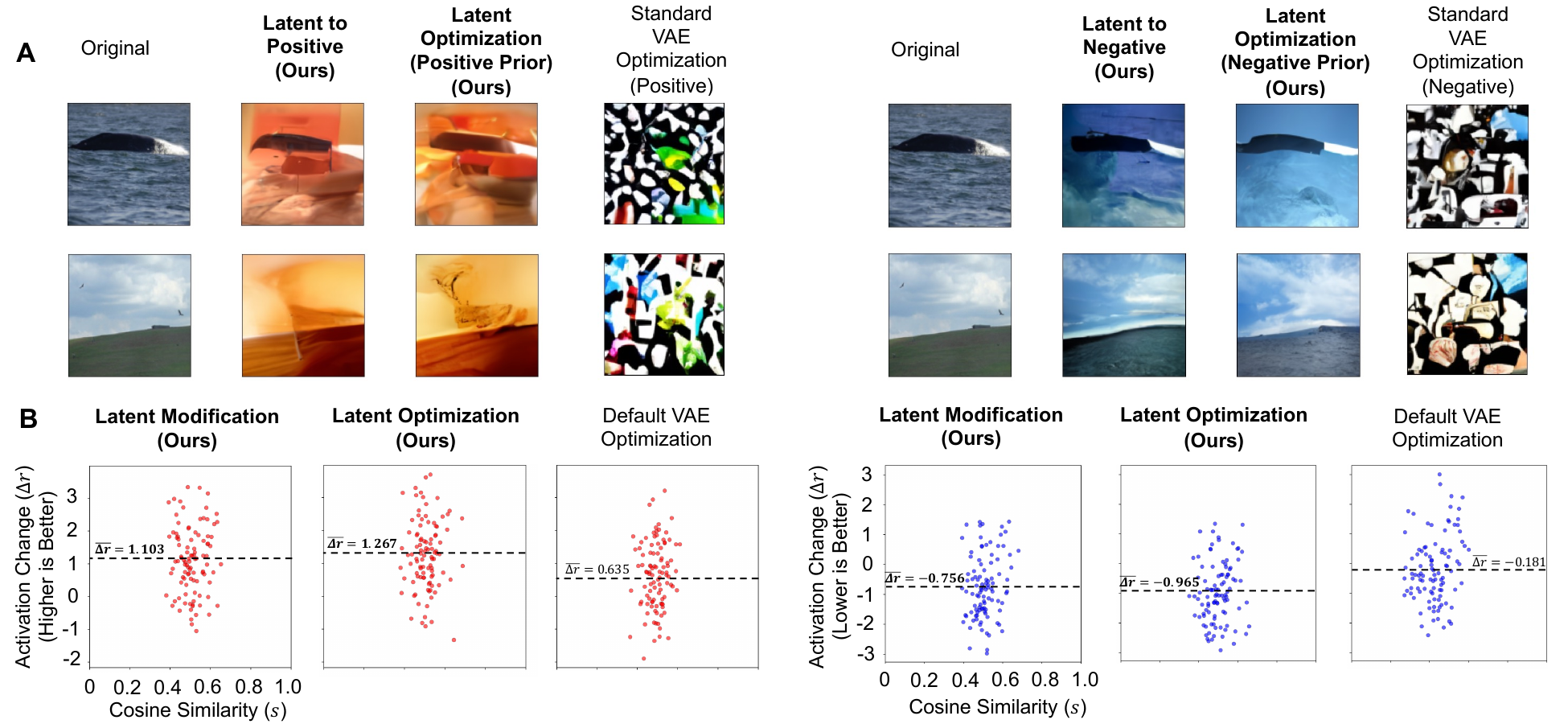}
   \caption{\textbf{a) Generating preferred images with different methods on ImageNet Dataset.} \textbf{Left 4 columns} show optimizing images to increase their activations. \textbf{Column 1} are four example input images. \textbf{Column 2-3} are results of our method while \textbf{column 4} represent optimizing the latent space with a standard VAE. \textbf{Right 4 columns} are results of optimization in the same methods presenting in the same order, with target of decreasing activations. \textbf{b)} shows the shift in the activation value between and the cosine similarity between randomly chosen 100 original images and their optimized results over different methods. \textbf{Compared with baseline VAE}, our method can modulate activation better as well as maintain similarity with original images, meaning unrelated features are preserved.}
   \label{fig:optimization_results}
\end{figure*}

Figure~\ref{fig:NSD_Result} and Table~\ref{tab:combined_brain_results} show results on the NSD dataset. Our method produces activation changes in the intended direction across all reported ROIs, whereas the BrainACTIV \cite{garciacerdas2025brainactiv} fails to do so in several cases, especially for activation decrease. In particular, our method yields a larger decrease in activation in all reported regions, supporting the use of an independent negative prior rather than simply reversing the maximal direction used in the BrainACTIV. For activation increase, the advantage of our method is most evident in less-studied or mixed-selective regions such as TE2p, IP0, and IP1, while performance is more comparable in strongly category-selective regions such as OFA, FFA, EBA, and OPA. Our method also consistently produces more realistic visual representations, as indicated by lower FID scores~\citep{FID}. Those results demonstrate that, compared with linear interpolation method in BrainACTIV, which the negative direction is simply the reverse of the positive one, the key advantage of our method is constructing a structured space to make in-distribution sampling easier, mitigating out-of-distribution problem in linear interpolation method, which could produce consistently correct activation change, and generating more realistic images because sample from such latent space will avoid over-shift from original image representations. This is particularly useful when the target representation is noisy or mixed-selective.

To examine the robustness of the predictor, we further compare activation modulation results obtained with the original subject-specific predictor and with a held-out-subject predictor on the NSD dataset. The results are shown in Table~\ref{tab:cross_subject}. The comparable trends across the two predictors suggest that our conclusions are not driven solely by a single subject-specific model.

\subsubsection{Generate Preferred Representations in Artificial Neural Networks}
\leavevmode\par
\label{subsubsec:Activate Unit in Artificial Neural Networks}
Besides biological targets, we also test our approach on purely \emph{artificial} targets: a mid-level unit inside a SimCLR encoder~\citep{chen2020simple}, a contrastive-learning framework with a ResNet-50 backbone, as well as a unit from a more complex Vision Transformer (ViT). Figure~\ref{fig:ANN_activation_manipulation} shows natural images that produce the largest responses from the chosen units in these ANNs. With our model trained on unit activations, editing the latent code toward the positive prior can generate representations that better activate the selected ANN units than the BrainACTIV \cite{garciacerdas2025brainactiv} in these examples.

\subsubsection{Optimize Unit Activations while Preserving Unrelated Features}
\label{subsubsec:Optimizing Unit Activations while Preserving Local Features}

\leavevmode\par
To demonstrate the strength of our structured VAE latent space, we trained our model with ImageNet dataset.
Standard VAE with a standard Gaussian prior. Then we constrain latent of VAE with \textbf{multiple units} activations, and only optimize the latent dimensions controlled by a certain unit. Compared to vanilla VAEs, our method achieves more significant changes in unit responses while preserving unrelated essential image characteristics. Besides modifying latent to prior values, which can manipulate activations and preserve features effectively, we further optimize the VAE's latent variables \( z \) using gradient-based methods, applying the Adam optimizer to adjust \( z \) such that the activation \( r \) of targeted network units is either maximized or minimized. The optimization objective combines the desired activation change with a KL divergence (KLD) regularization term, ensuring that the latent representation remains aligned with the prior distribution. The results of latent modification and optimization on both VAEs are shown in Figure~\ref{fig:optimization_results}.

To further validate the effectiveness of our optimization approach. Cosine similarity, computed using features extracted from a pre-trained \textbf{ResNet18} \cite{he2016deep} encoder, quantifies the preservation of image features. Results in Figure~\ref{fig:optimization_results} indicate that our structured landscape approach not only achieves more robust activation changes but also maintains similarity to the original images compared to standard VAE methods, underscoring the efficacy of incorporating structured priors for latent optimization.
\section{Discussion and Limitations}
\label{sec:Discussion and Limitations}
We developed \modelname{}, a framework that imposes an activation-regularized prior on a VAE to yield a structured latent landscape and then performs direct latent editing to steer activations in biological and artificial targets. This structured landscape can also modulate the activation of target units while preserving unrelated features. \modelname{} shows clear advantages in changing activations in correct direction, keeping image realism and decreasing activations under our evaluation protocol. Inhibit activation is  meaningful in clinical applications, reducing abnormally elevated activity in specific circuits has been associated with therapeutic benefit. For example, prior work has shown that attenuation of amygdala hyper-reactivity is linked to improvement in anxiety and trauma-related symptoms \cite{Inhi1}, while reduction of subgenual cingulate hyperactivity has been associated with antidepressant treatment effects \cite{Inhi2}. Thus, although our study is entirely in-silico and does not constitute a medical intervention, the ability to selectively suppress target-unit responses may be relevant for modeling clinically meaningful forms of neural modulation.

Our framework can also extend to artificial neural networks as a controlled setting for studying feature visualization. Compared with the linear method, \modelname{} does not assume that activation increase and decrease follow a single linear direction in CLIP space. Instead, it learns activation-specific regions in a structured latent space, and enables more controllable sampling, which is especially useful when neural responses are noisy and mixed-selective.

There are also limitations to our work. The image generator and the CLIP encoder are frozen, so biases inherited from their pre-training data may propagate into the generated preferred images. %Our evaluation is entirely in silico, meaning that the generated images are assessed through pretrained response predictors rather than closed-loop biological experiments.

In the future, we may scale the framework to additional modalities such as EEG, MEG or larger-scale brain datasets, and move beyond in-silico evaluation by testing the generated preferred images in human experiments; such closed-loop validation would further confirm the effectiveness of our framework and broaden its potential use in brain-computer-interface applications.

\bibliographystyle{ACM-Reference-Format}
\bibliography{main}
% Supplementary material starts on a new page, in one column for tables and algorithms.
\clearpage
\onecolumn
\appendix
\renewcommand{\thesection}{\Alph{section}}

\renewcommand{\thefigure}{S\arabic{figure}}
\renewcommand{\thetable}{S\arabic{table}}

\renewcommand{\theHfigure}{supp.\arabic{figure}}
\renewcommand{\theHtable}{supp.\arabic{table}}
\setcounter{figure}{0}
\setcounter{table}{0}

\begin{center}
  {\LARGE\bfseries Supplementary Material\par}
\end{center}
\vspace{0.5em}
\subsection*{Appendix Overview}

This appendix provides comprehensive details and additional results to support the findings presented in the main paper and ensure that all of the results can be reproduced. The appendix content is organized into the following sections:

\begin{itemize}
    \item \textbf{Additional Results}: Additional visualizations and analyses that further demonstrate the effectiveness of our proposed methods. More examples of enhanced sampling, local change results are presented, as shown in section~\ref{sec:Additional Results}.

    \item \textbf{Experimental Settings}: Comprehensive information about the datasets settings, like input size and how are the datasets divided and processed, model architectures and implementation details, then training configurations and computational resources. All of them can provide a clear understanding of our experimental setup, which are helpful to reproduce the results, as shown in section~\ref{sec:Experimental Settings}.

    \item \textbf{Implementation Details}: Detailed information about the algorithms we used to implement our method, including pseudocode, and corresponding explanations. All of them can provide a clear understanding of how the training and optimization are carried out, as shown in section~\ref{sec:Implementation Details}.

    \item \textbf{Predictive Power of In-silico brain simulator}: Information about how much our trained In-silico brain simulator can predict the brain activity data, as shown in section~\ref{sec:Predictive Power}.
\end{itemize}

\newcommand{\logvar}{\mathrm{logvar}}

\section{Additional Results}
\label{sec:Additional Results}
\subsection{More Results for the NSD Dataset}

Here is a summarized table of more experimental results on NSD dataset:

\begin{table}[h]
\centering
\small
\caption{Activation change and realism on additional ROIs. Better values are in \textbf{bold}.}
\renewcommand{\arraystretch}{1.15}
\resizebox{1.0\linewidth}{!}{
\begin{tabular}{llcccccccc}
\toprule
        & ROIs & V1 & V2 & V3 & V4 & V8 & LO1 & LO2 & V4t\\
\midrule
\multirow{2}{*}{\makecell[l]{\textbf{Activation}\\[-2pt]\textbf{Increase} ($\Delta r$)}}
        & Ours      & \textbf{0.257} & 0.080 & \textbf{0.247} & 0.486 & 0.696 & \textbf{0.583} & 0.498 & 0.757 \\
        & BrainACTIV  & 0.157 & \textbf{0.116} & 0.226 & \textbf{0.488} & \textbf{0.721} & 0.582 & \textbf{0.517} & \textbf{0.767}\\
\midrule
\multirow{2}{*}{\makecell[l]{\textbf{Activation}\\[-2pt]\textbf{Decrease} ($\Delta r$)}}
        & Ours      & \textbf{-0.291} & \textbf{-0.272} & \textbf{-0.573} & \textbf{-0.690} & \textbf{-0.406} & \textbf{-0.485} & \textbf{-0.490} & \textbf{-0.601}\\
        & BrainACTIV  & 0.102 & 0.224 & 0.274 & 0.501 & 0.608 & 0.100 & -0.057 & -0.511\\
\midrule
\multirow{2}{*}{\textbf{FID@2k}}
        & Ours      & \textbf{164.87} & \textbf{157.84} & \textbf{167.44} & \textbf{173.24} & \textbf{160.07} & \textbf{136.05} & \textbf{186.67} & \textbf{178.62}\\
        & BrainACTIV  & 224.29 & 201.56 & 240.64 & 231.60 & 216.80 & 208.23 & 255.06 & 197.61\\
\bottomrule
\end{tabular}
}
\vspace{0.7em}% -------------- small gap between blocks --------------

\resizebox{1.0\linewidth}{!}{
\begin{tabular}{llcccccccc}
\toprule
        & ROIs & VWFA-1 & PPA & RSC & OWFA & IP2 & TPOJ1 & TPOJ2 & TPOJ3\\
\midrule
\multirow{2}{*}{\makecell[l]{\textbf{Activation}\\[-2pt]\textbf{Increase} ($\Delta r$)}}
        & Ours      & \textbf{1.108} & \textbf{0.952} & 0.923 & 0.718 & \textbf{0.091} & 0.456 & 0.661 & 1.014\\
        & BrainACTIV  & 1.019 & 0.926 & \textbf{0.954} & \textbf{1.088} & 0.084 & \textbf{0.509} &\textbf{0.751} & \textbf{1.091}\\
\midrule
\multirow{2}{*}{\makecell[l]{\textbf{Activation}\\[-2pt]\textbf{Decrease} ($\Delta r$)}}
        & Ours      & \textbf{-0.391} & \textbf{-0.809} & \textbf{-0.913} & \textbf{-0.780} & \textbf{-0.213} & \textbf{-0.136} & \textbf{-0.384} & \textbf{-0.591}\\
        & BrainACTIV  & 0.350 & -0.579 & -0.804 & 0.347 & 0.061 & -0.134  & -0.234 & -0.484\\
\midrule
\multirow{2}{*}{\textbf{FID@2k}}
        & Ours      & \textbf{169.14} & \textbf{153.67} & \textbf{172.92} & \textbf{163.88} & \textbf{168.56} & \textbf{192.04} & \textbf{131.83} & \textbf{137.57}\\
        & BrainACTIV  & 229.65 & 234.79 & 216.65 & 207.15 & 171.94 & 244.44 & 197.94 & 197.24\\
\bottomrule
\end{tabular}
}
\label{tab:ex_ROI_appendix}
\end{table}

The full test-set results for the additional ROIs are provided in Figs. S1–S16.

\clearpage
\subsubsection{ROI: IPS4}
\leavevmode\par
\nopagebreak[4]
\begin{figure}[H]
    \centering
    \includegraphics[width=0.95\textwidth,height=0.90\textheight,keepaspectratio]{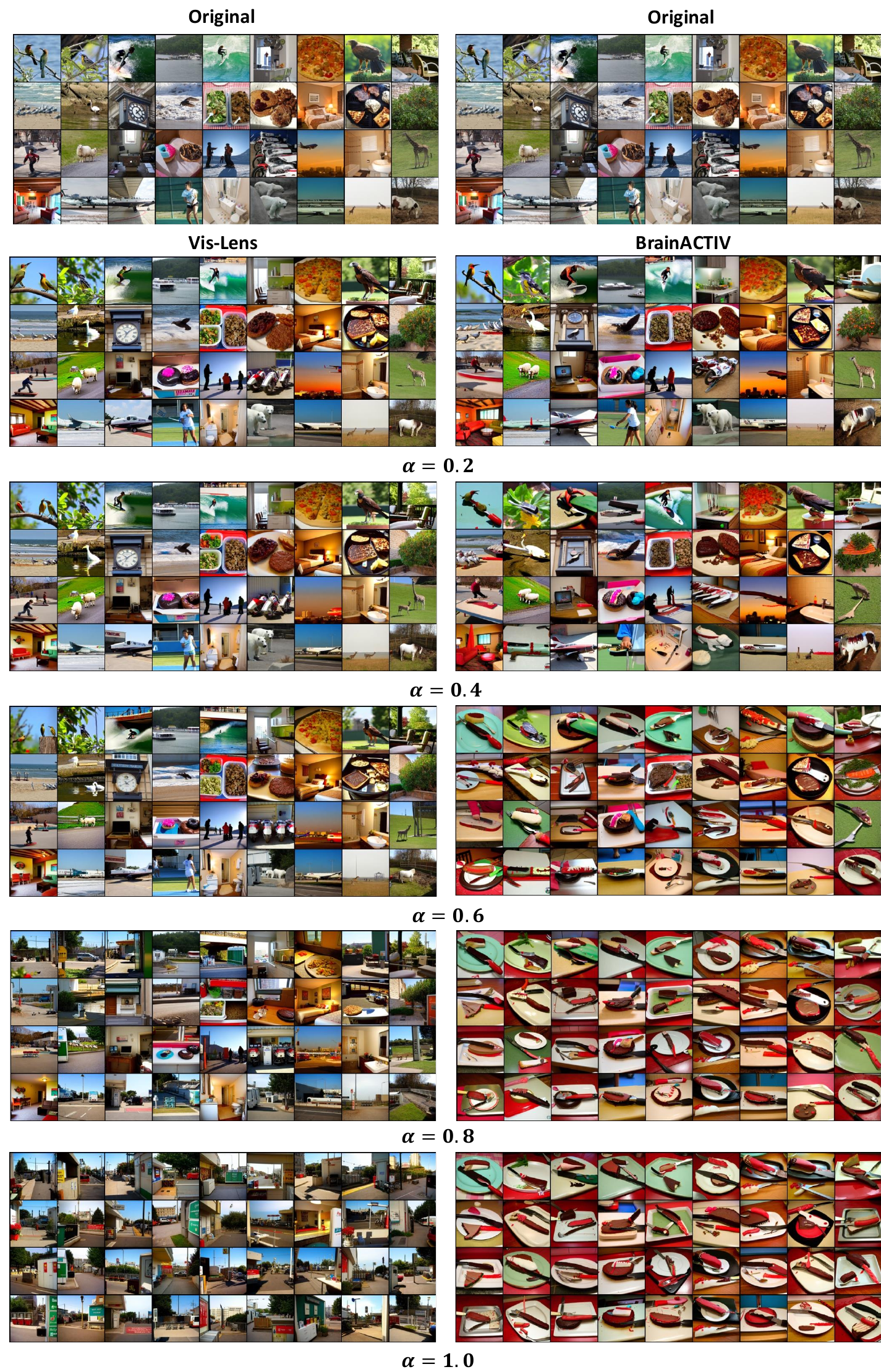}
    \caption{Full result of activation change on region IPS4, positive direction.} 
    \label{fig:nsd_1_pos}
\end{figure}

\clearpage
\begin{figure}[H]
    \centering
    \includegraphics[width=0.95\textwidth,height=0.90\textheight,keepaspectratio]{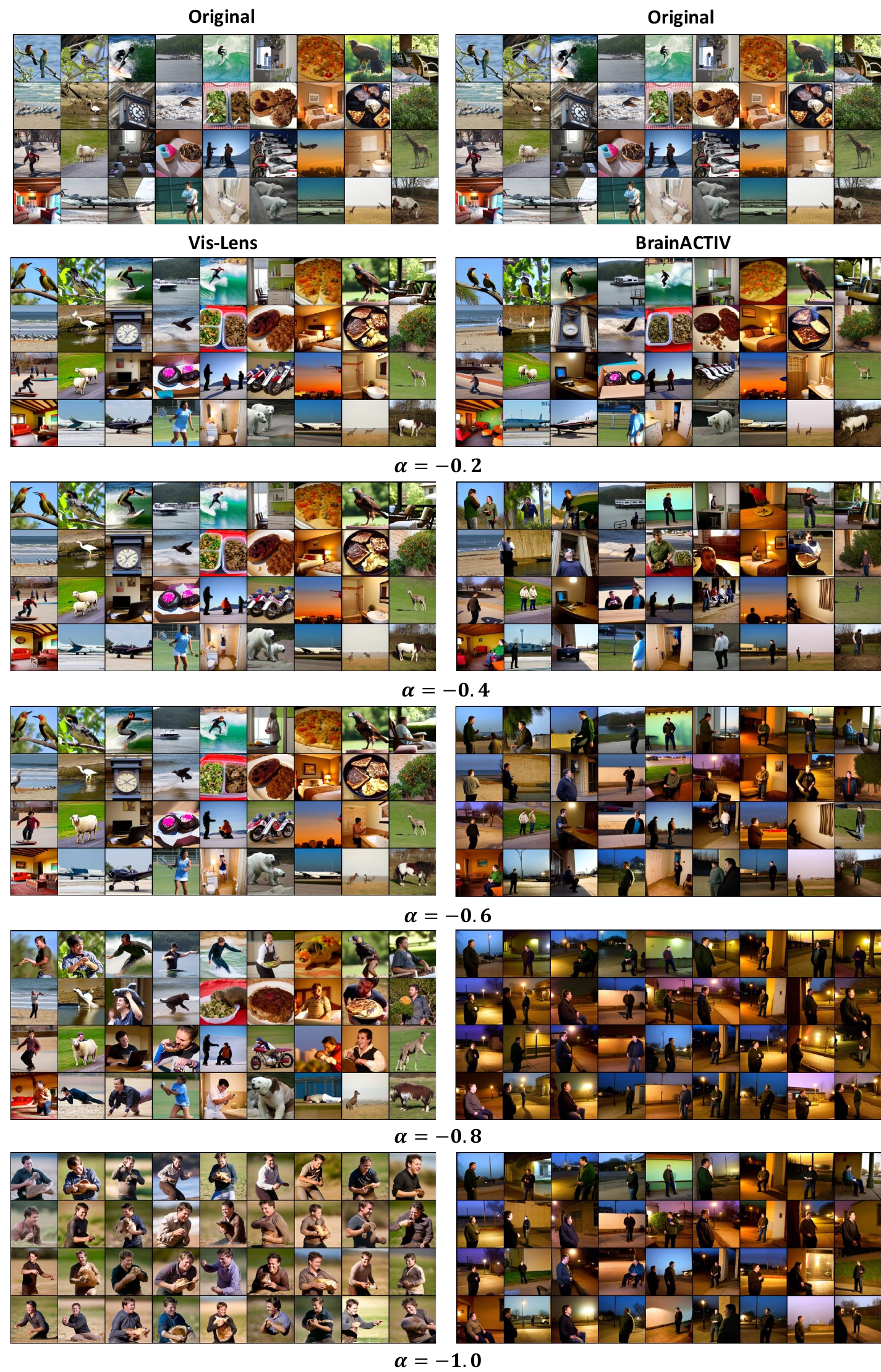}
    \caption{Full result of activation change on region IPS4, negative direction.} 
    \label{fig:nsd_1_neg}
\end{figure}

\clearpage
\subsubsection{ROI: SPL1}
\leavevmode\par
\nopagebreak[4]
\begin{figure}[H]
    \centering
    \includegraphics[width=0.95\textwidth,height=0.90\textheight,keepaspectratio]{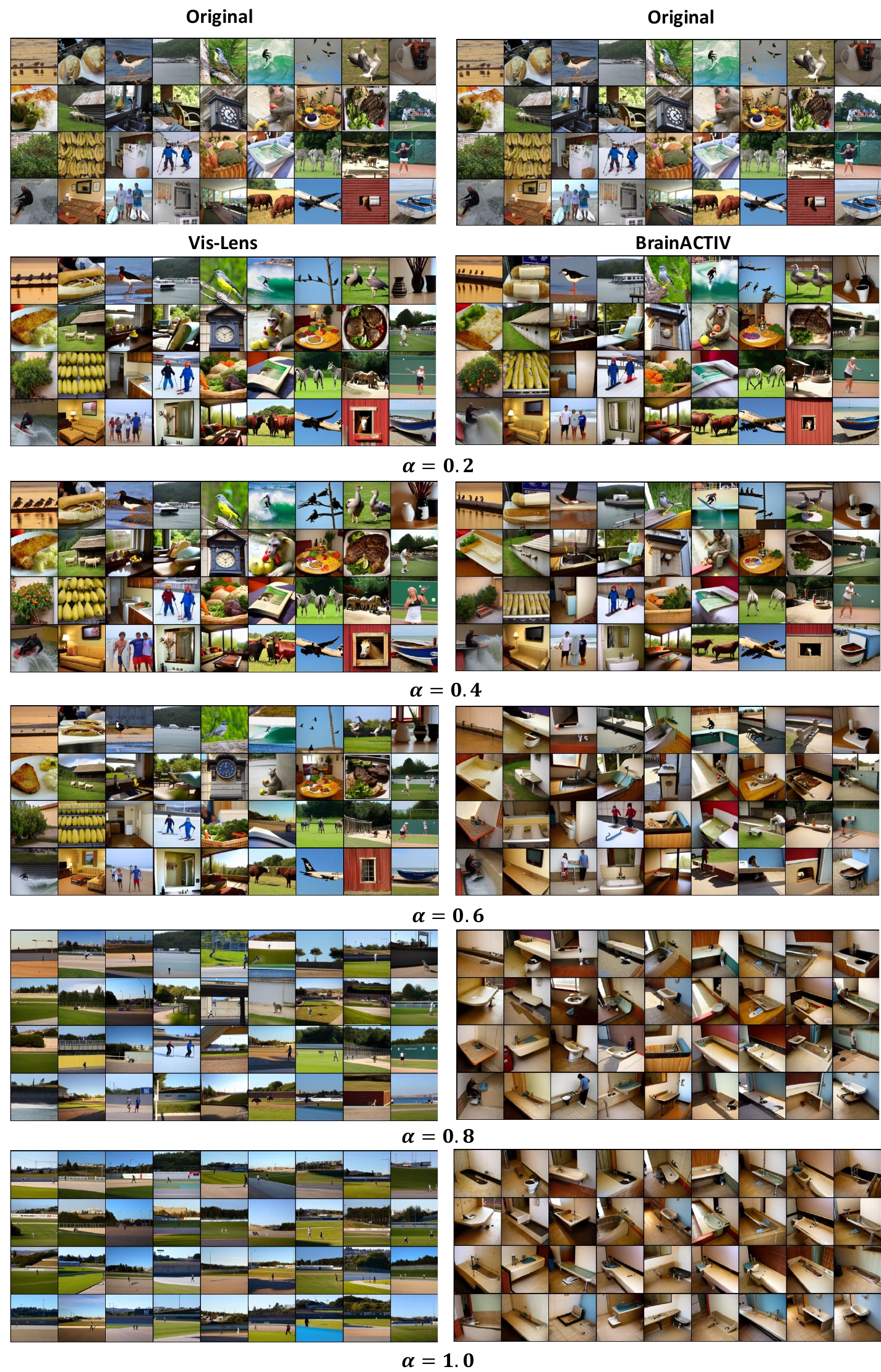}
    \caption{Full result of activation change on region SPL1, positive direction.} 
    \label{fig:nsd_2_pos}
\end{figure}

\clearpage
\begin{figure}[H]
    \centering
    \includegraphics[width=0.95\textwidth,height=0.90\textheight,keepaspectratio]{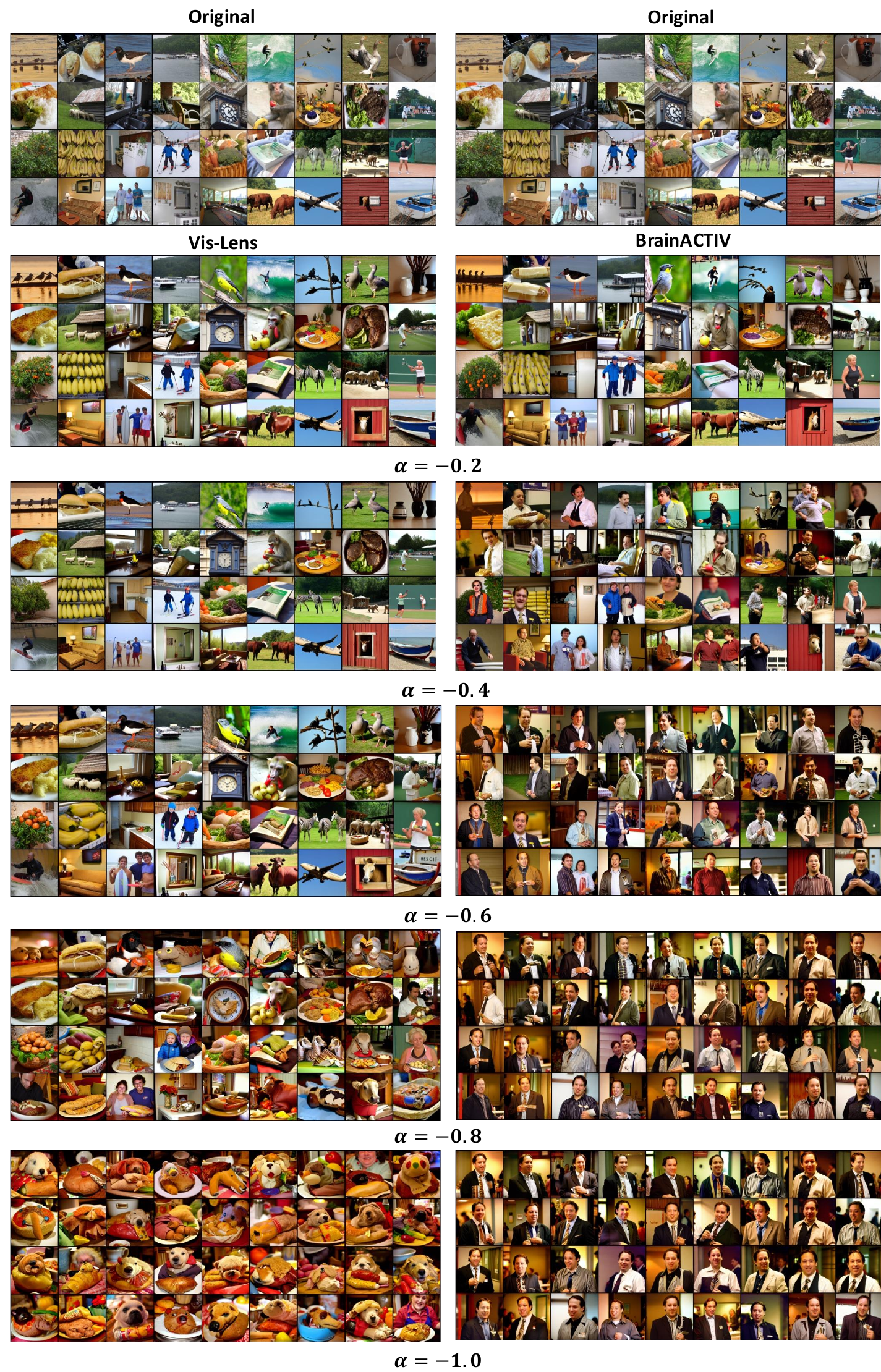}
    \caption{Full result of activation change on region SPL1, negative direction.} 
    \label{fig:nsd_2_neg}
\end{figure}

\clearpage
\subsubsection{ROI: 31a}
\leavevmode\par
\nopagebreak[4]
\begin{figure}[H]
    \centering
    \includegraphics[width=0.95\textwidth,height=0.90\textheight,keepaspectratio]{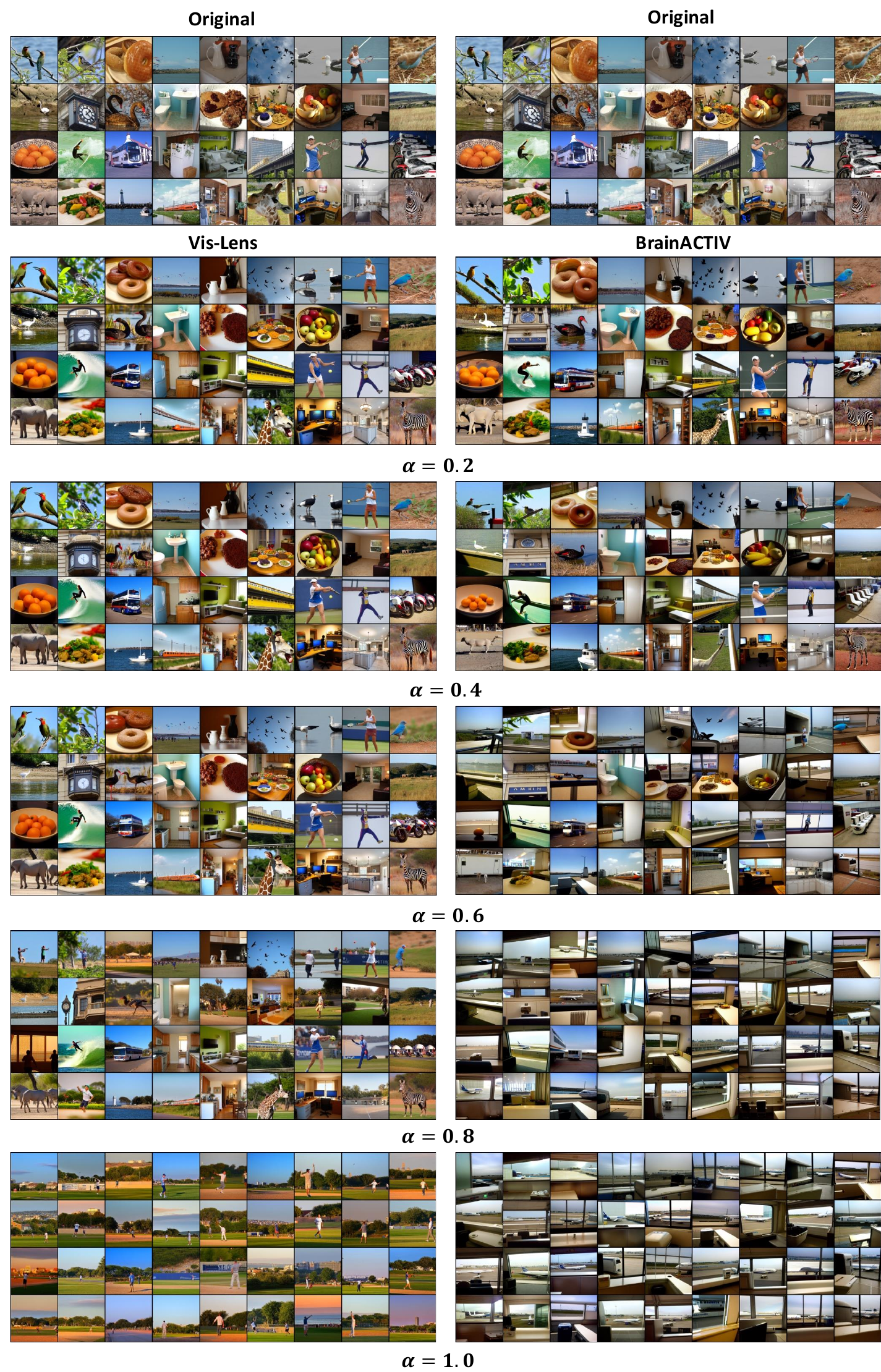}
    \caption{Full result of activation change on region 31a, positive direction.} 
    \label{fig:nsd_3_pos}
\end{figure}

\clearpage
\begin{figure}[H]
    \centering
    \includegraphics[width=0.95\textwidth,height=0.90\textheight,keepaspectratio]{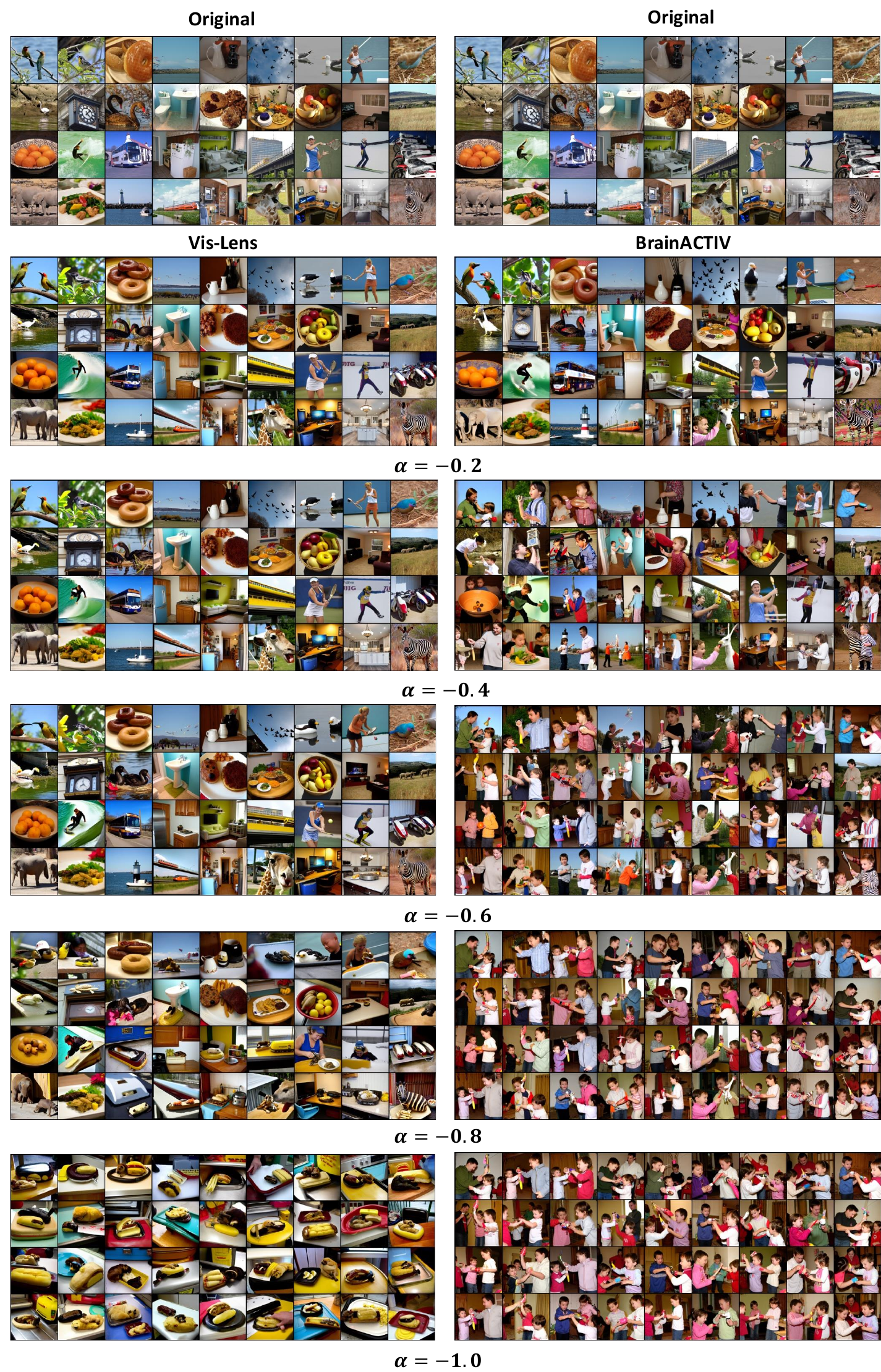}
    \caption{Full result of activation change on region 31a, negative direction.} 
    \label{fig:nsd_3_neg}
\end{figure}

\clearpage
\subsubsection{ROI: 31pd}
\leavevmode\par
\nopagebreak[4]
\begin{figure}[H]
    \centering
    \includegraphics[width=0.95\textwidth,height=0.90\textheight,keepaspectratio]{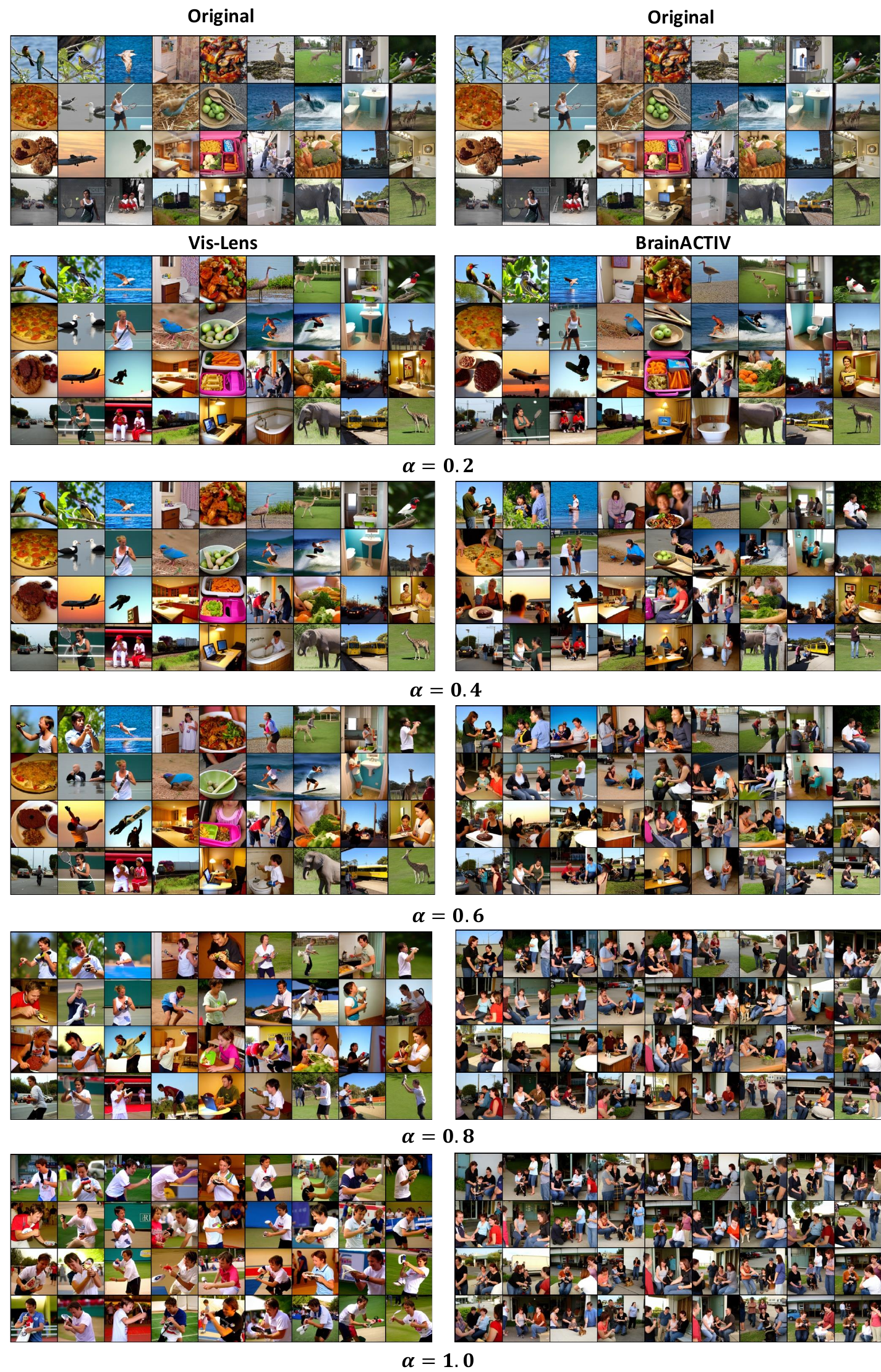}
    \caption{Full result of activation change on region 31pd, positive direction.} 
    \label{fig:nsd_4_pos}
\end{figure}

\clearpage
\begin{figure}[H]
    \centering
    \includegraphics[width=0.95\textwidth,height=0.90\textheight,keepaspectratio]{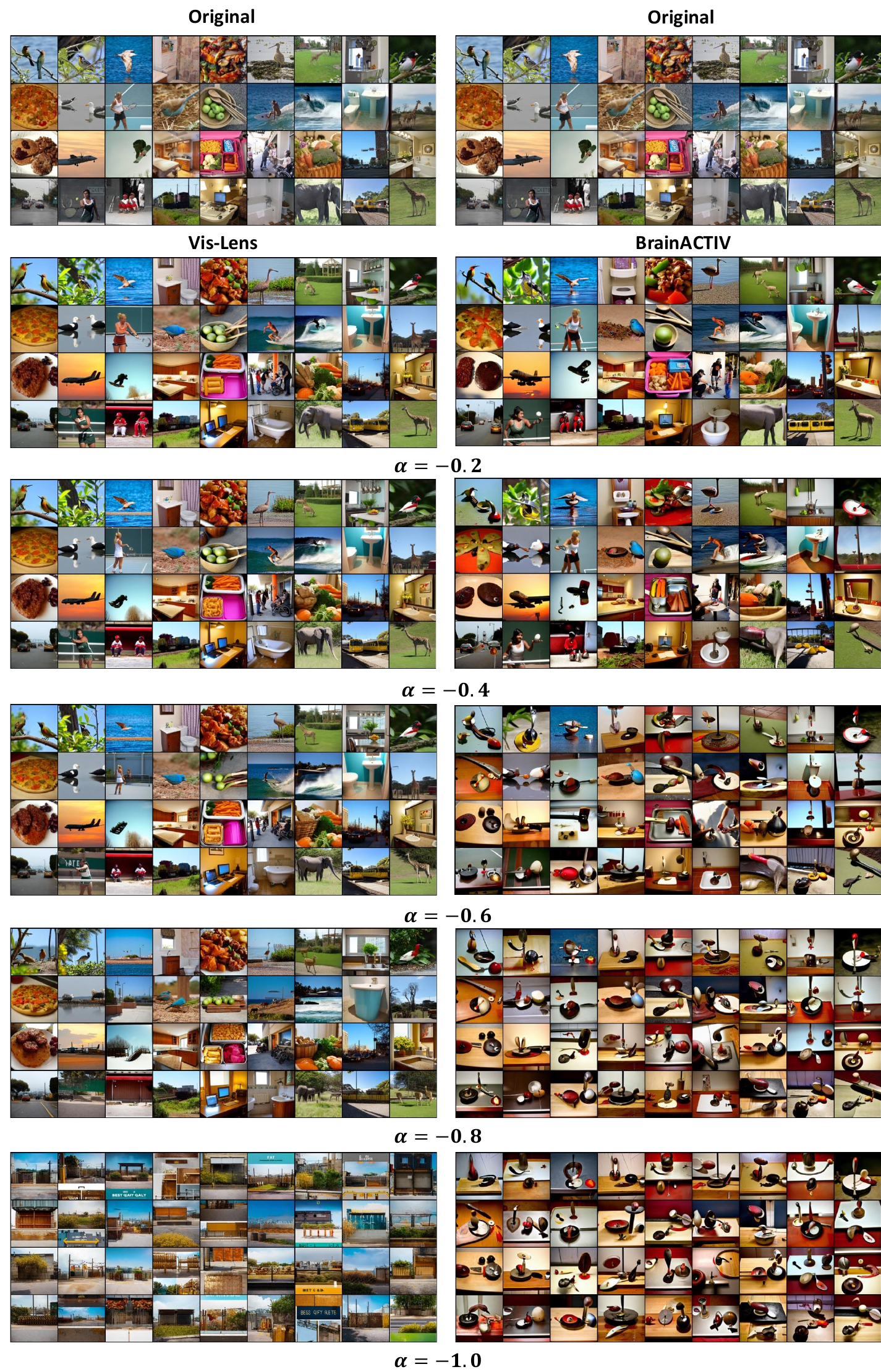}
    \caption{Full result of activation change on region 31pd, negative direction.} 
    \label{fig:nsd_4_neg}
\end{figure}

\clearpage
\subsubsection{ROI: V1}
\leavevmode\par
\nopagebreak[4]
\begin{figure}[H]
    \centering
    \includegraphics[width=0.95\textwidth,height=0.90\textheight,keepaspectratio]{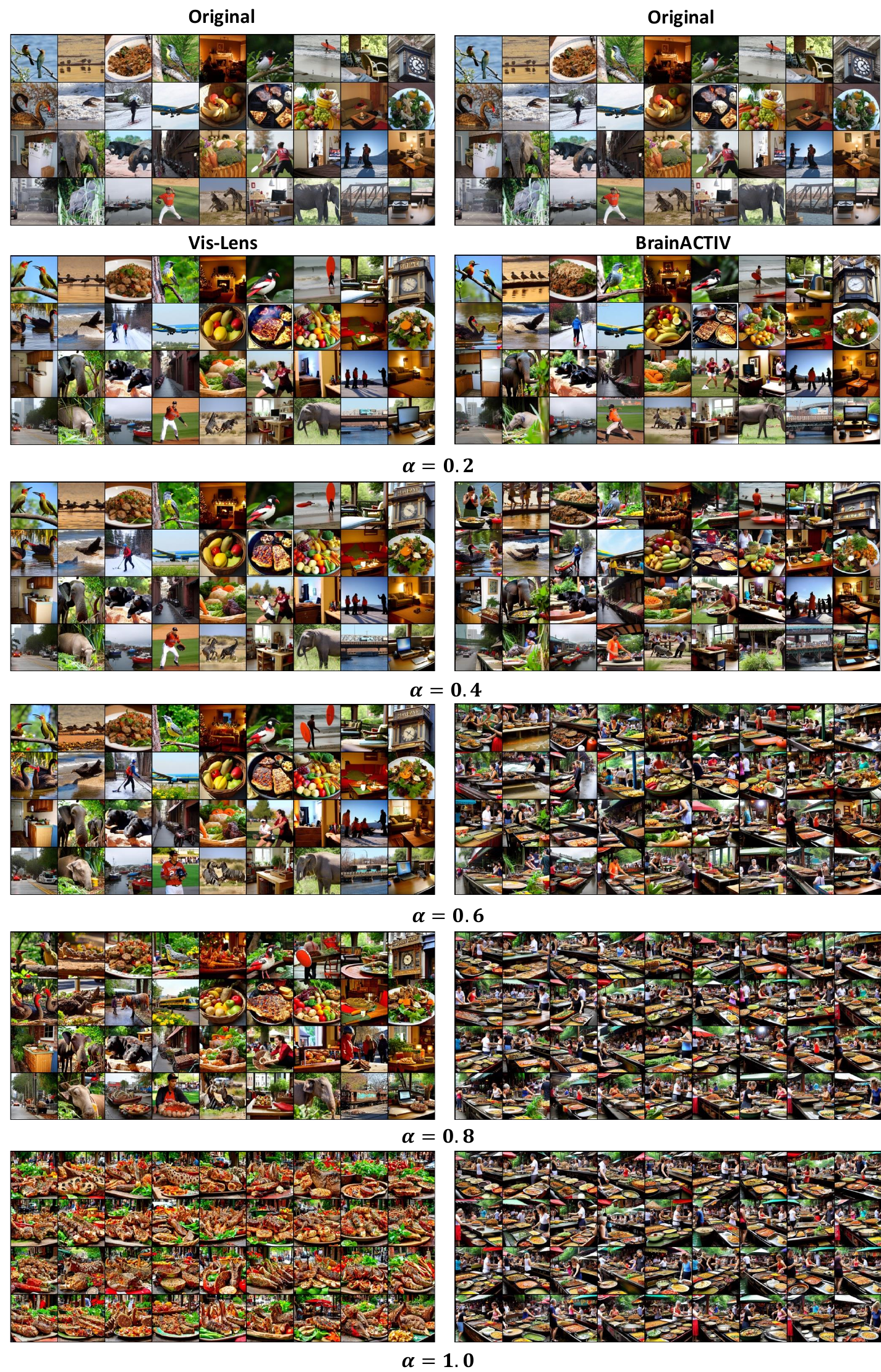}
    \caption{Full result of activation change on region V1, positive direction.} 
    \label{fig:nsd_5_pos}
\end{figure}

\clearpage
\begin{figure}[H]
    \centering
    \includegraphics[width=0.95\textwidth,height=0.90\textheight,keepaspectratio]{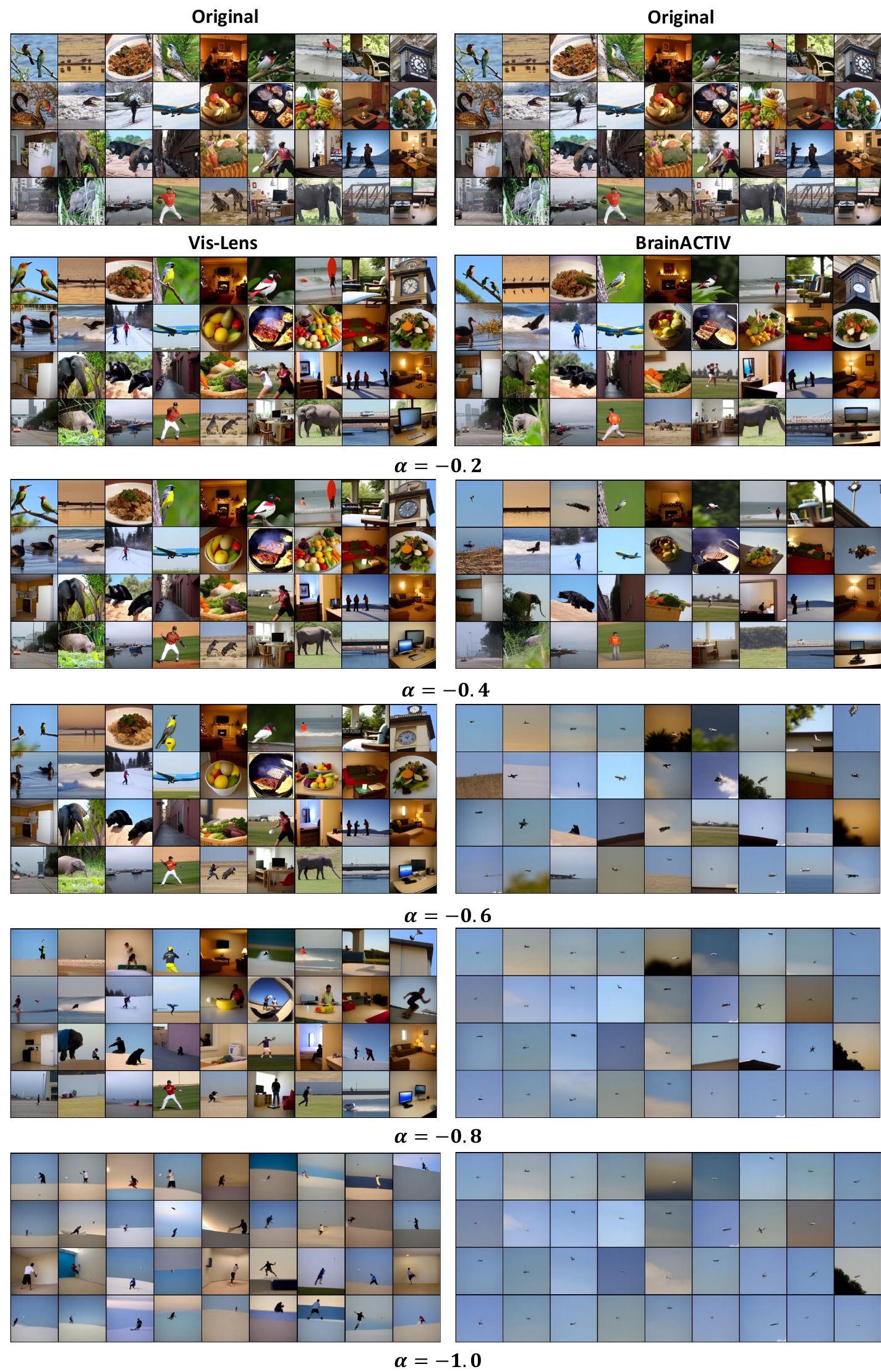}
    \caption{Full result of activation change on region V1, negative direction.} 
    \label{fig:nsd_5_neg}
\end{figure}

\clearpage
\subsubsection{ROI: V2}
\leavevmode\par
\nopagebreak[4]
\begin{figure}[H]
    \centering
    \includegraphics[width=0.95\textwidth,height=0.90\textheight,keepaspectratio]{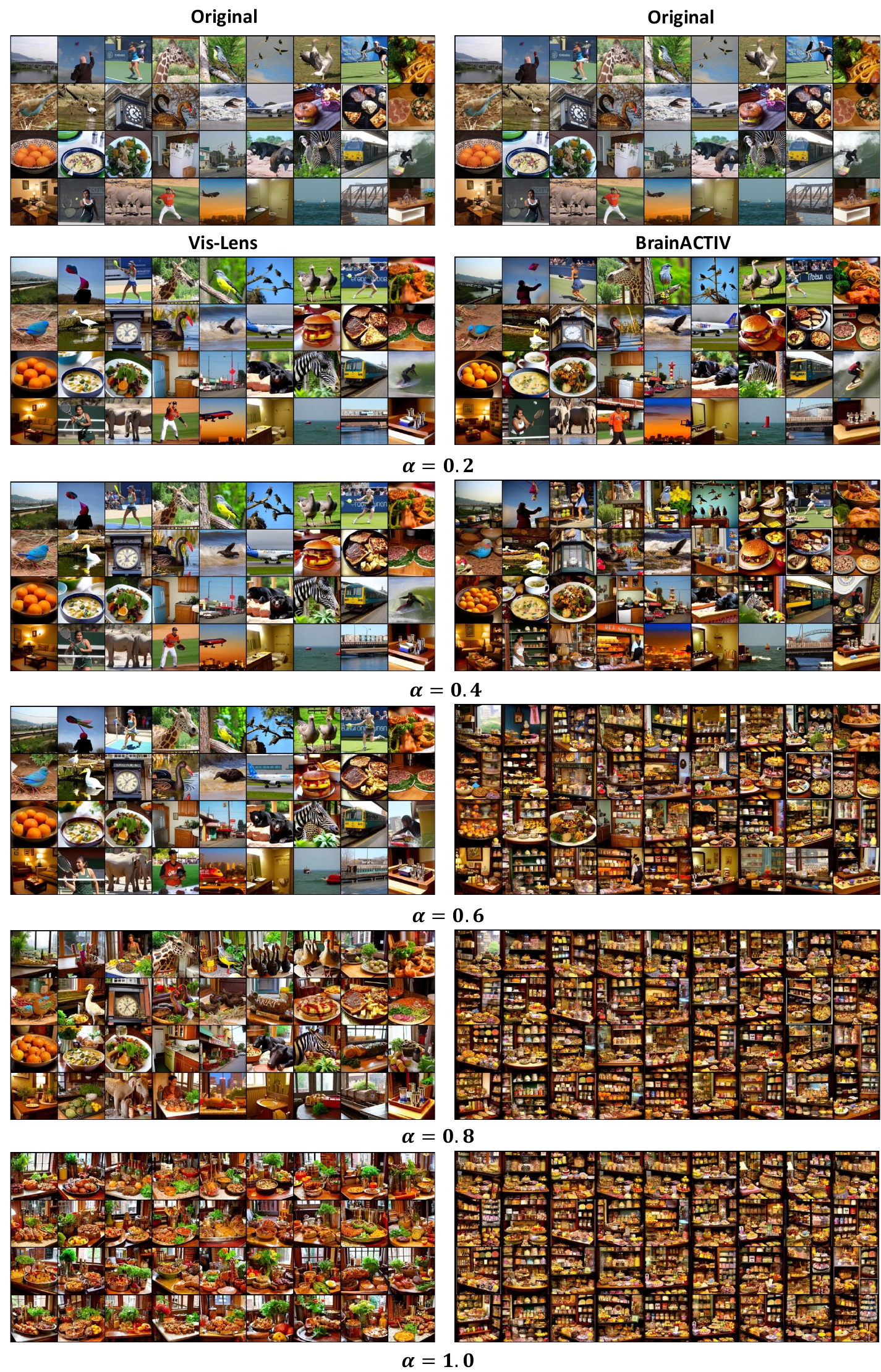}
    \caption{Full result of activation change on region V2, positive direction.} 
    \label{fig:nsd_6_pos}
\end{figure}

\clearpage
\begin{figure}[H]
    \centering
    \includegraphics[width=0.95\textwidth,height=0.90\textheight,keepaspectratio]{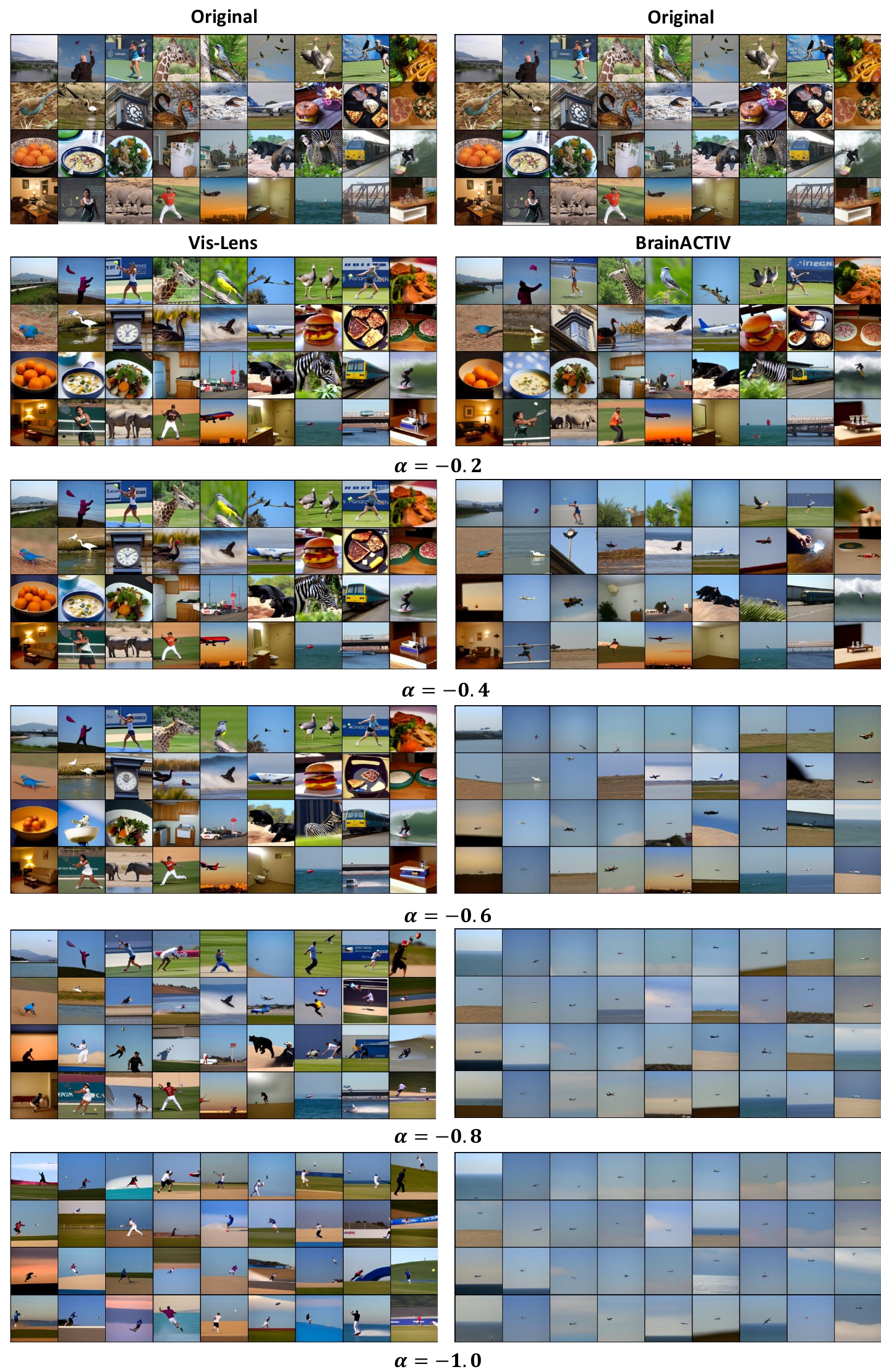}
    \caption{Full result of activation change on region V2, negative direction.} 
    \label{fig:nsd_6_neg}
\end{figure}

\clearpage
\subsubsection{ROI: V3}
\leavevmode\par
\nopagebreak[4]
\begin{figure}[H]
    \centering
    \includegraphics[width=0.95\textwidth,height=0.90\textheight,keepaspectratio]{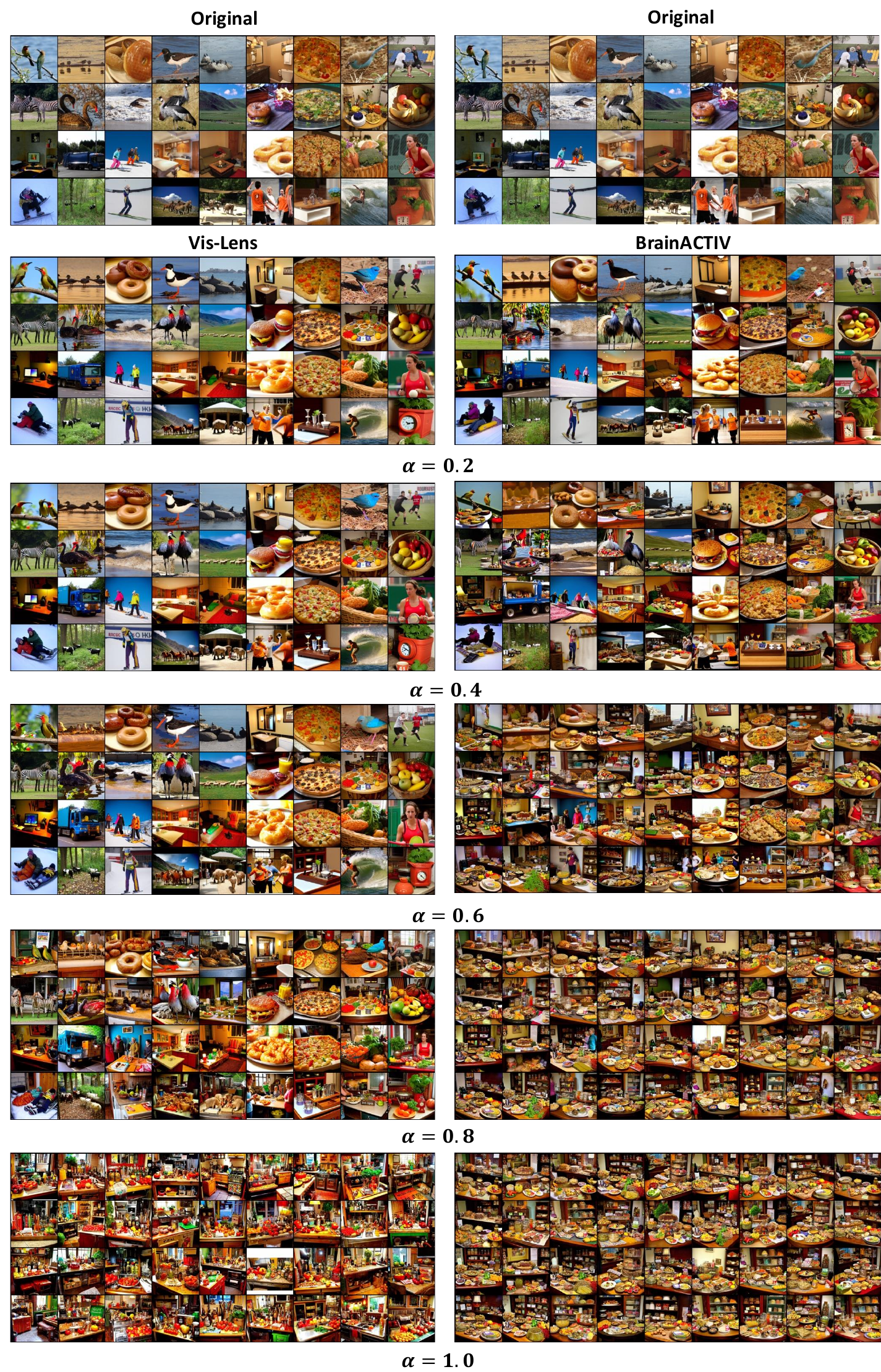}
    \caption{Full result of activation change on region V3, positive direction.} 
    \label{fig:nsd_7_pos}
\end{figure}

\clearpage
\begin{figure}[H]
    \centering
    \includegraphics[width=0.95\textwidth,height=0.90\textheight,keepaspectratio]{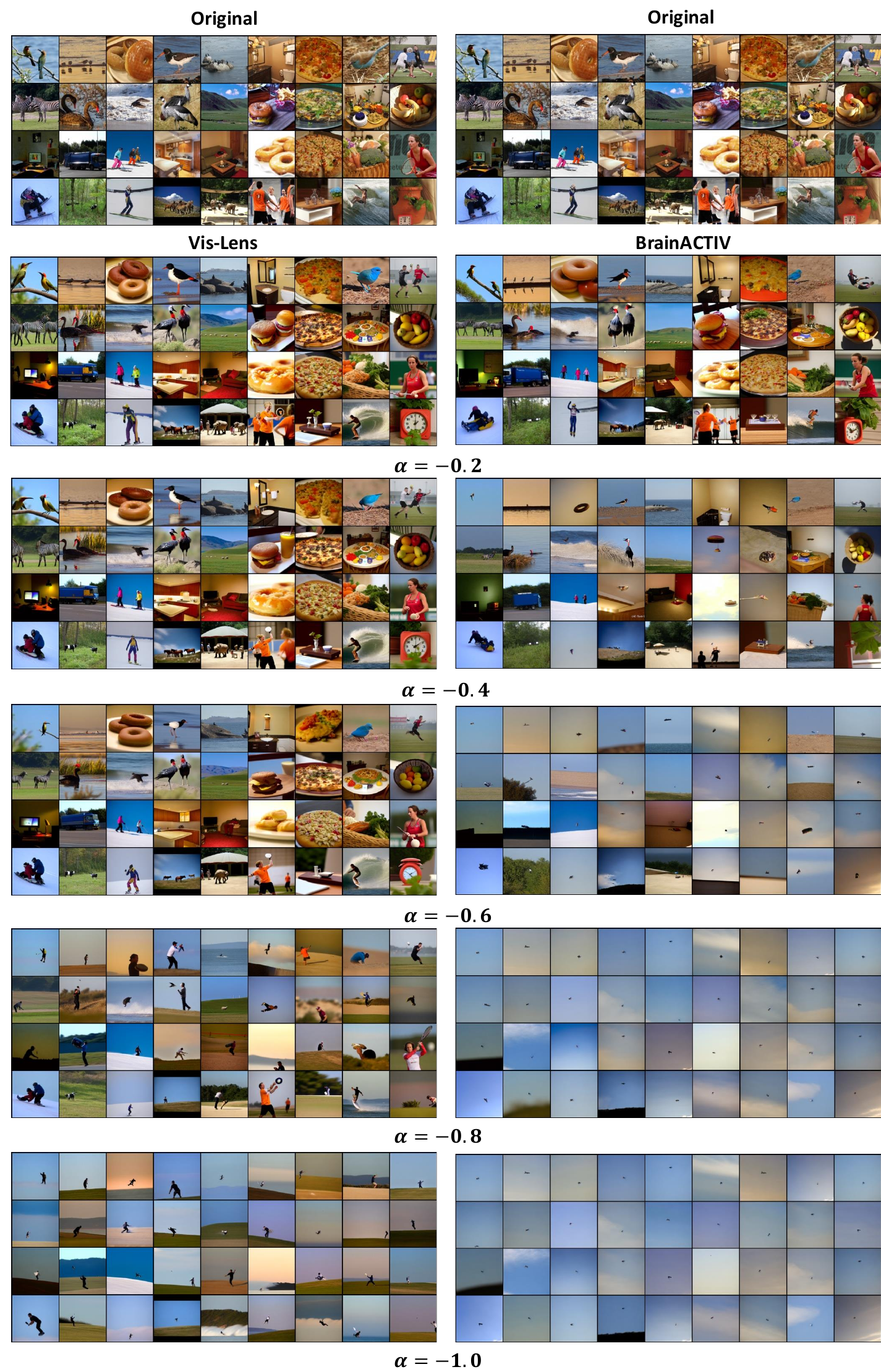}
    \caption{Full result of activation change on region V3, negative direction.} 
    \label{fig:nsd_7_neg}
\end{figure}

\clearpage
\subsubsection{ROI: V4}
\leavevmode\par
\nopagebreak[4]
\begin{figure}[H]
    \centering
    \includegraphics[width=0.95\textwidth,height=0.90\textheight,keepaspectratio]{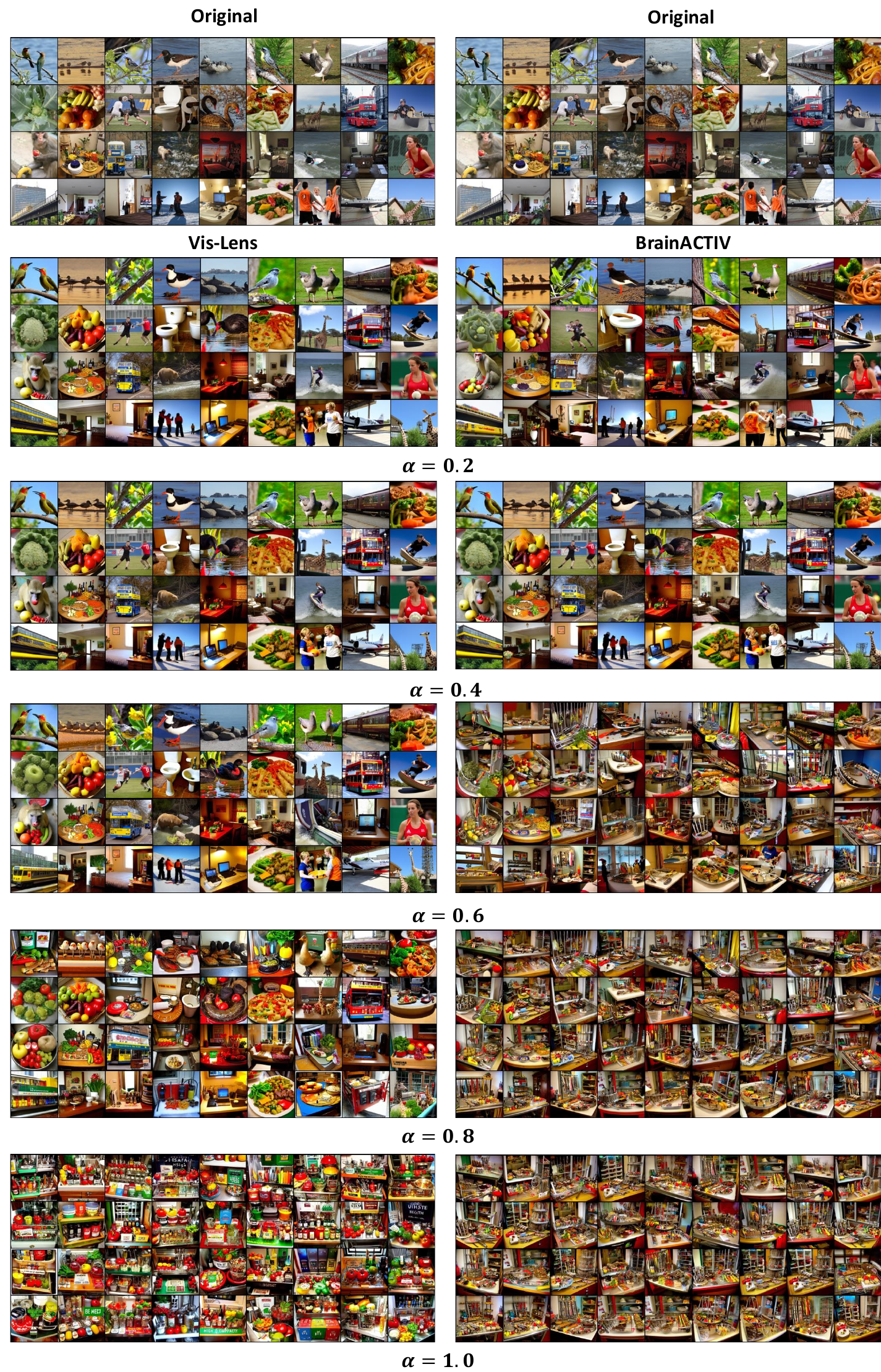}
    \caption{Full result of activation change on region V4, positive direction.} 
    \label{fig:nsd_8_pos}
\end{figure}

\clearpage
\begin{figure}[H]
    \centering
    \includegraphics[width=0.95\textwidth,height=0.90\textheight,keepaspectratio]{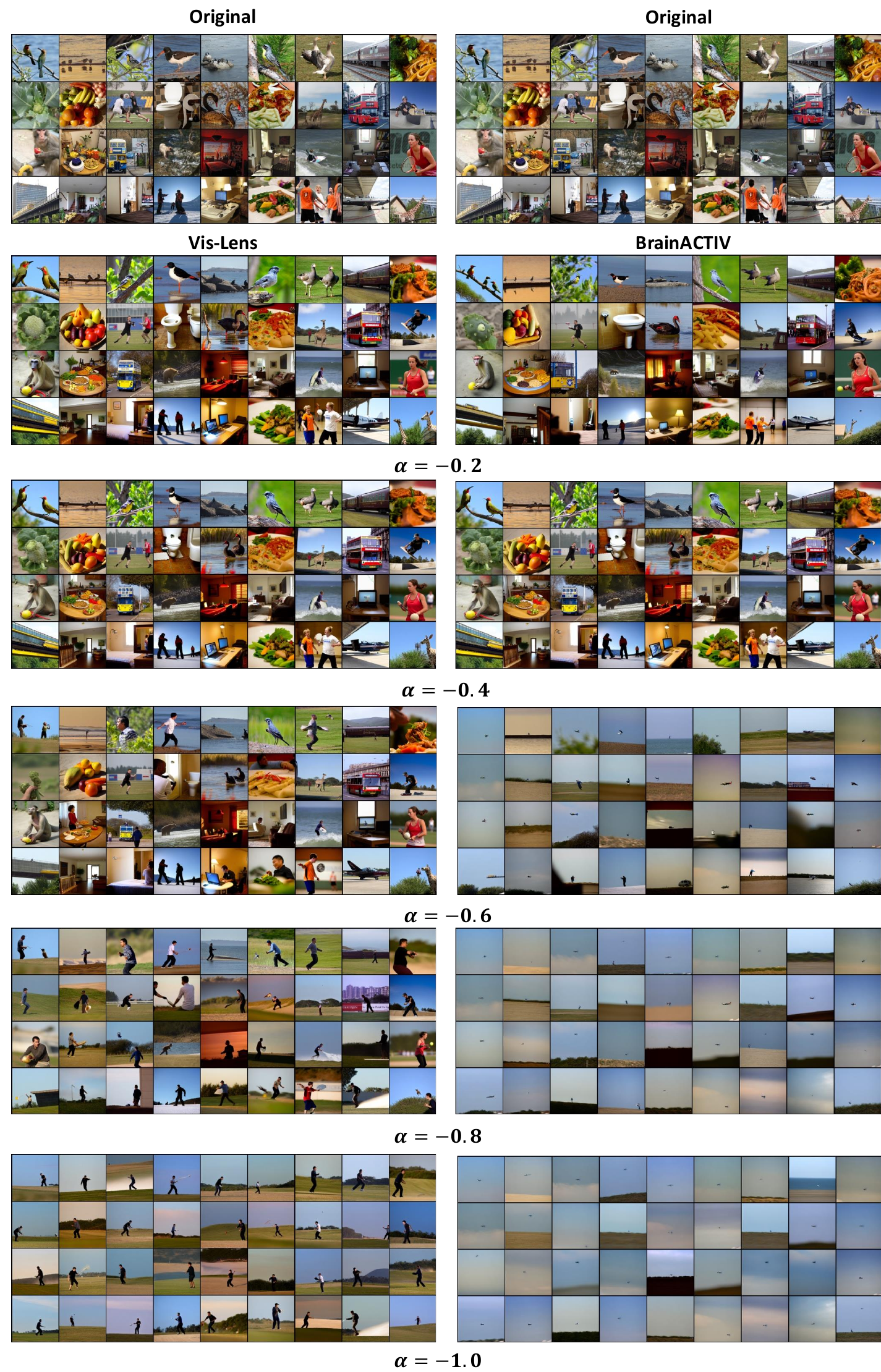}
    \caption{Full result of activation change on region V4, negative direction.} 
    \label{fig:nsd_8_neg}
\end{figure}

\clearpage
\subsection{More ANN Results}
We present more examples on manipulating activations in SimCLR\citep{chen2020simple}, shown in Figs. S17–S22, which further demonstrate the effectiveness of our method.

\begin{figure}[H]
  \centering
  \includegraphics[width=\linewidth]{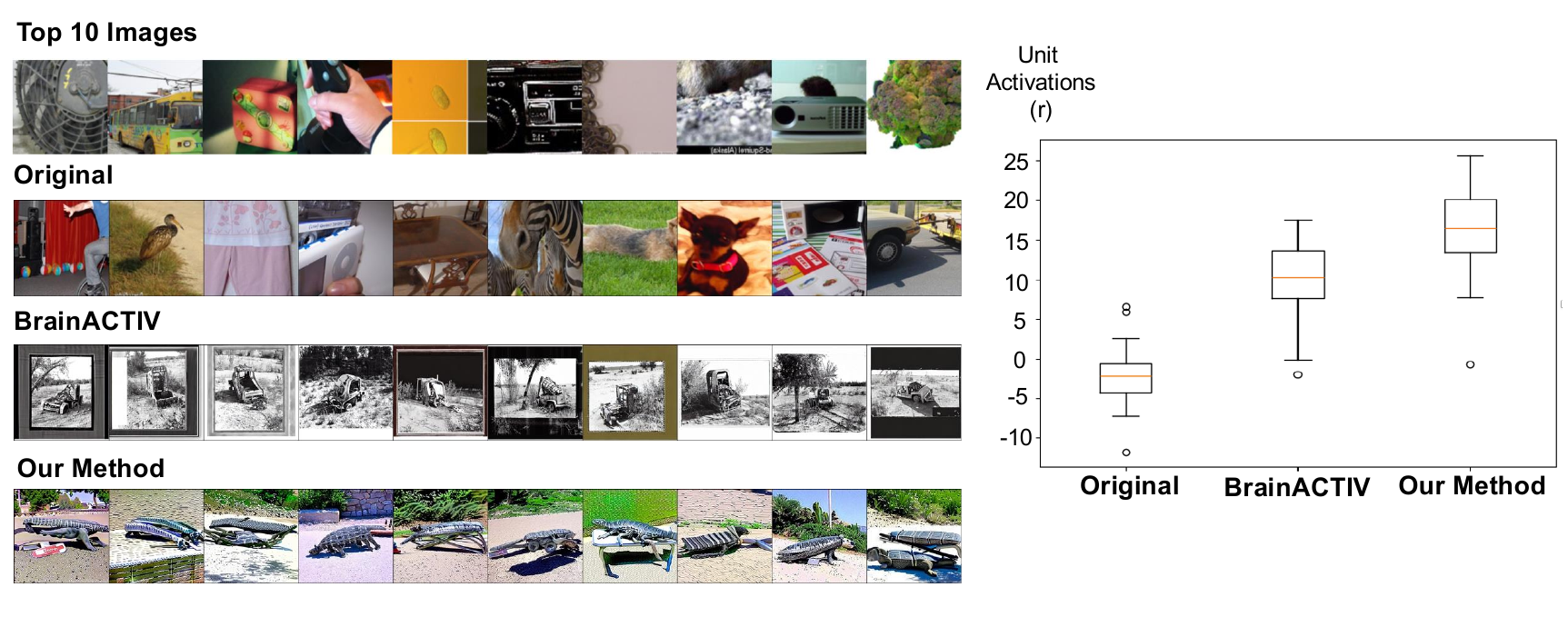}
   \caption{\textbf{One latent modification example to generate preferred images with increasing unit activations on ImageNet-mini Dataset.} Though the top 10 images can not demonstrate a notable pattern represented by the unit in the early layer of SimCLR model, our method can generate a uniform representation and compared with original images from ImageNet-mini, generated preferred images based on modifying the latent to the positive prior mean shows a more apparent change in activation compared with original images and representations from BrainACTIV.}
   \label{fig:ImageNet_2}
\end{figure}

\begin{figure}[H]
  \centering
  \includegraphics[width=1\textwidth]{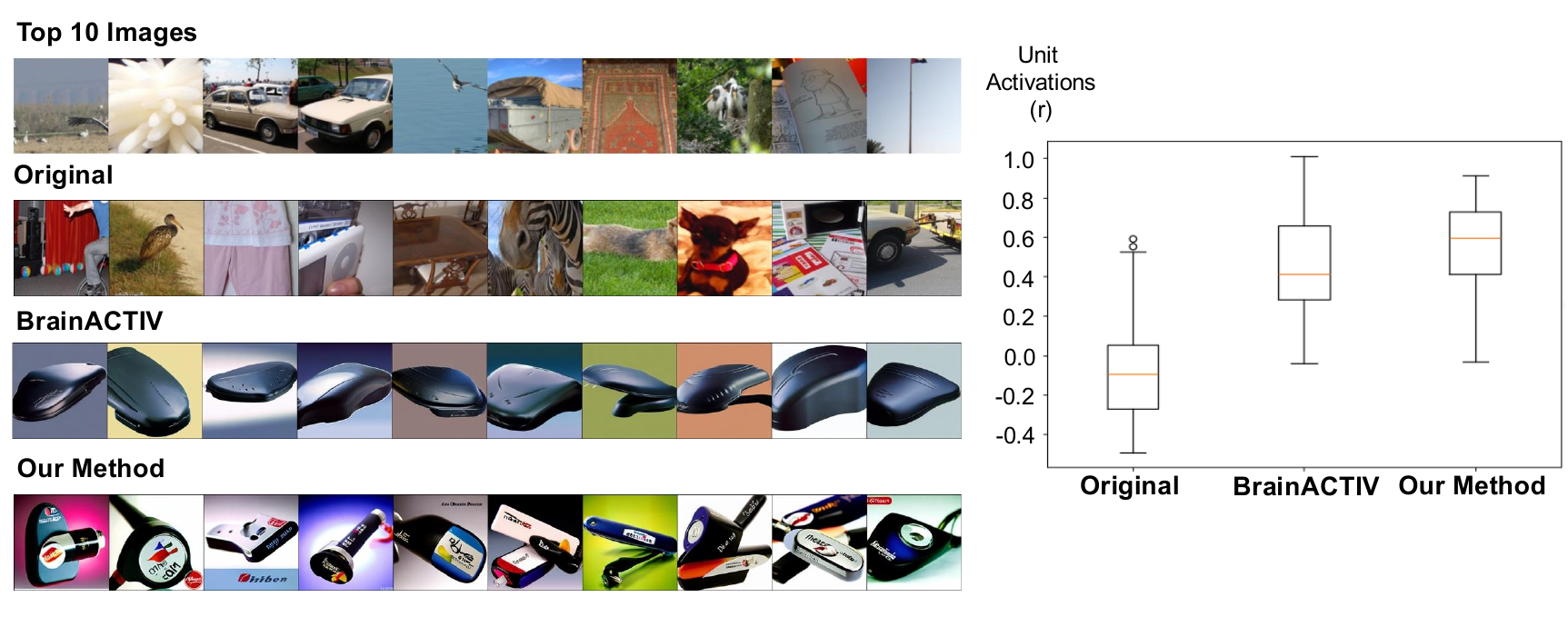}
   \caption{\textbf{One latent modification example to generate preferred images with increasing unit activations on ImageNet-mini Dataset.} Though the top 10 images can not demonstrate a notable pattern represented by the unit in the early layer of SimCLR model, our method can generate a uniform representation and compared with original images from ImageNet-mini, generated preferred images based on modifying the latent to the positive prior mean shows a more apparent change in activation compared with original images and representations from BrainACTIV.}
   \label{fig:ImageNet_3}
\end{figure}

\begin{figure}[H]
  \centering
  \includegraphics[width=\linewidth, height=0.35\textheight]{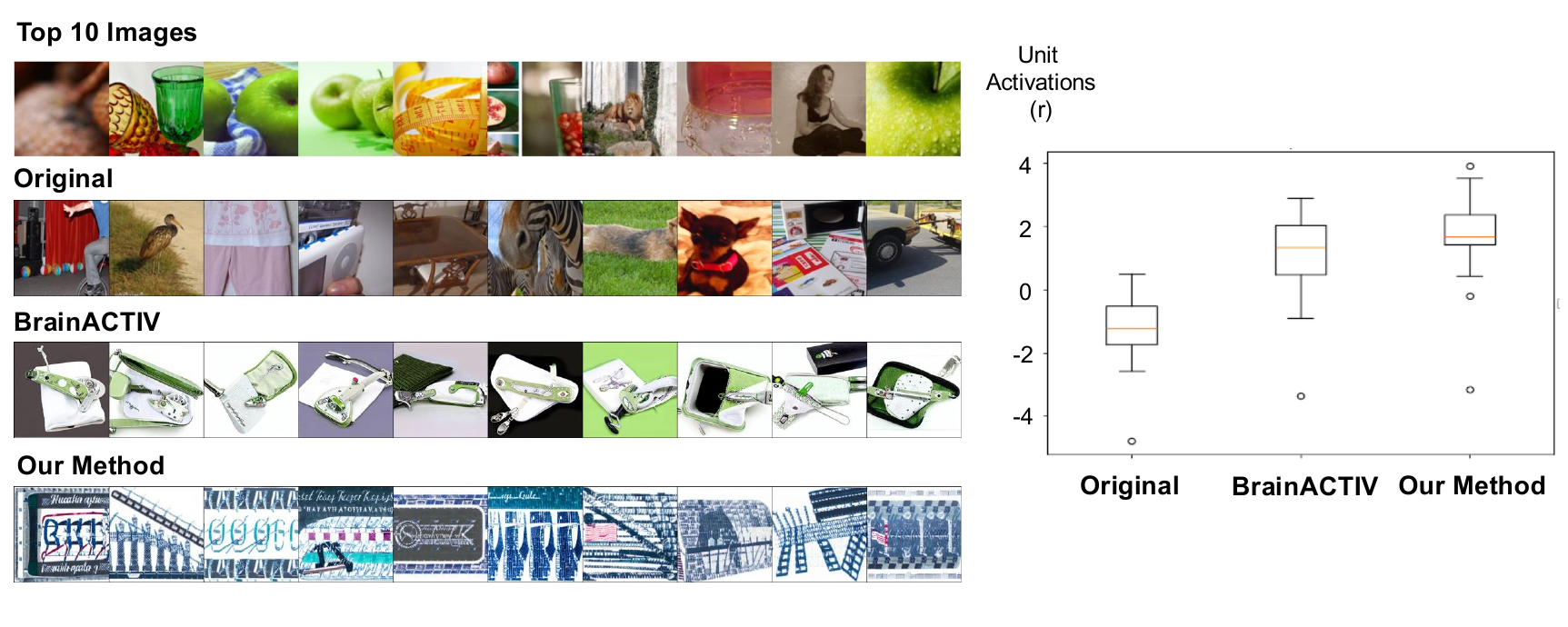}
   \caption{\textbf{One latent modification example to generate preferred images with increasing unit activations on ImageNet-mini Dataset.} Though the top 10 images can not demonstrate a notable pattern represented by the unit in the early layer of SimCLR model, our method can generate a uniform representation and compared with original images from ImageNet-mini, generated preferred images based on modifying the latent to the positive prior mean shows a more apparent change in activation compared with original images and representations from BrainACTIV.}
   \label{fig:ImageNet_4}
\end{figure}

\begin{figure}[H]
  \centering
  \includegraphics[width=\linewidth, height=0.35\textheight]{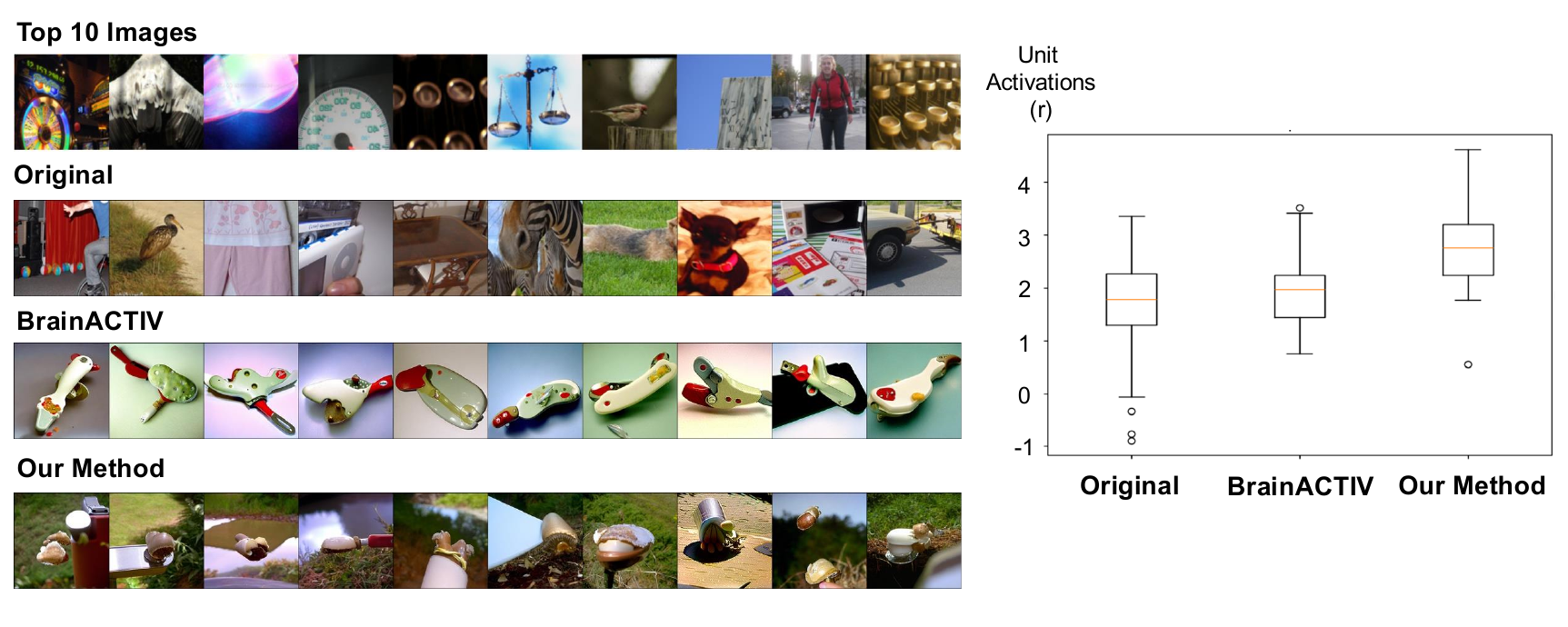}
   \caption{\textbf{One latent modification example to generate preferred images with increasing unit activations on ImageNet-mini Dataset.} The top 10 images demonstrate a notable pattern represented by the unit in the deep layer of SimCLR model, that round objects appear in top images. Our method can also generate a uniform representation and compared with original images from ImageNet-mini, generated preferred images based on modifying the latent to the positive prior mean shows a more apparent change in activation compared with original images and representations from BrainACTIV.}
   \label{fig:ImageNet_5}
\end{figure}

\begin{figure}[H]
  \centering
  \includegraphics[width=1\textwidth]{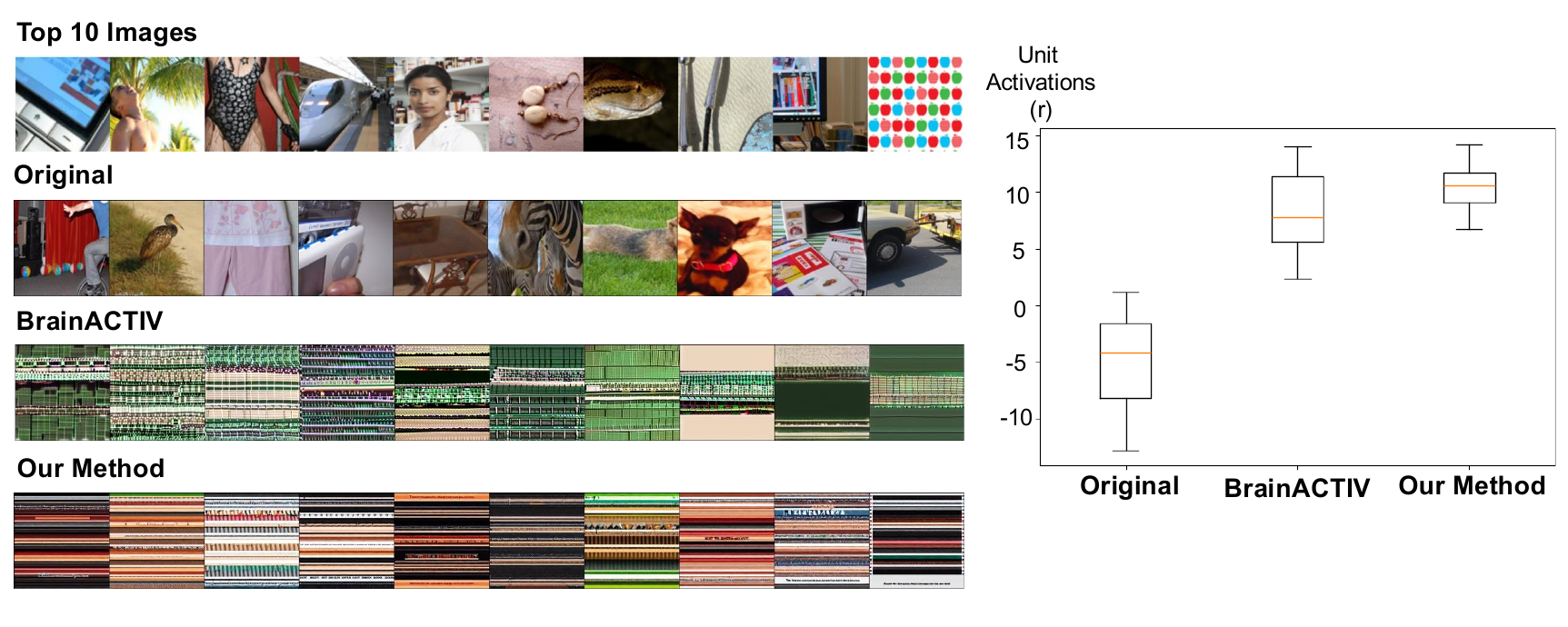}
   \caption{\textbf{One latent modification example to generate preferred images with increasing unit activations on ImageNet-mini Dataset.} The top 10 images demonstrate a notable pattern represented by the unit in the deep layer of SimCLR model, that grid objects appear in top images. Our method can also generate a uniform representation and compared with original images from ImageNet-mini, generated preferred images based on modifying the latent to the positive prior mean shows a more apparent change in activation compared with original images and representations from BrainACTIV.}
   \label{fig:ImageNet_6}
\end{figure}

\begin{figure}[H]
  \centering
  \includegraphics[width=1\textwidth]{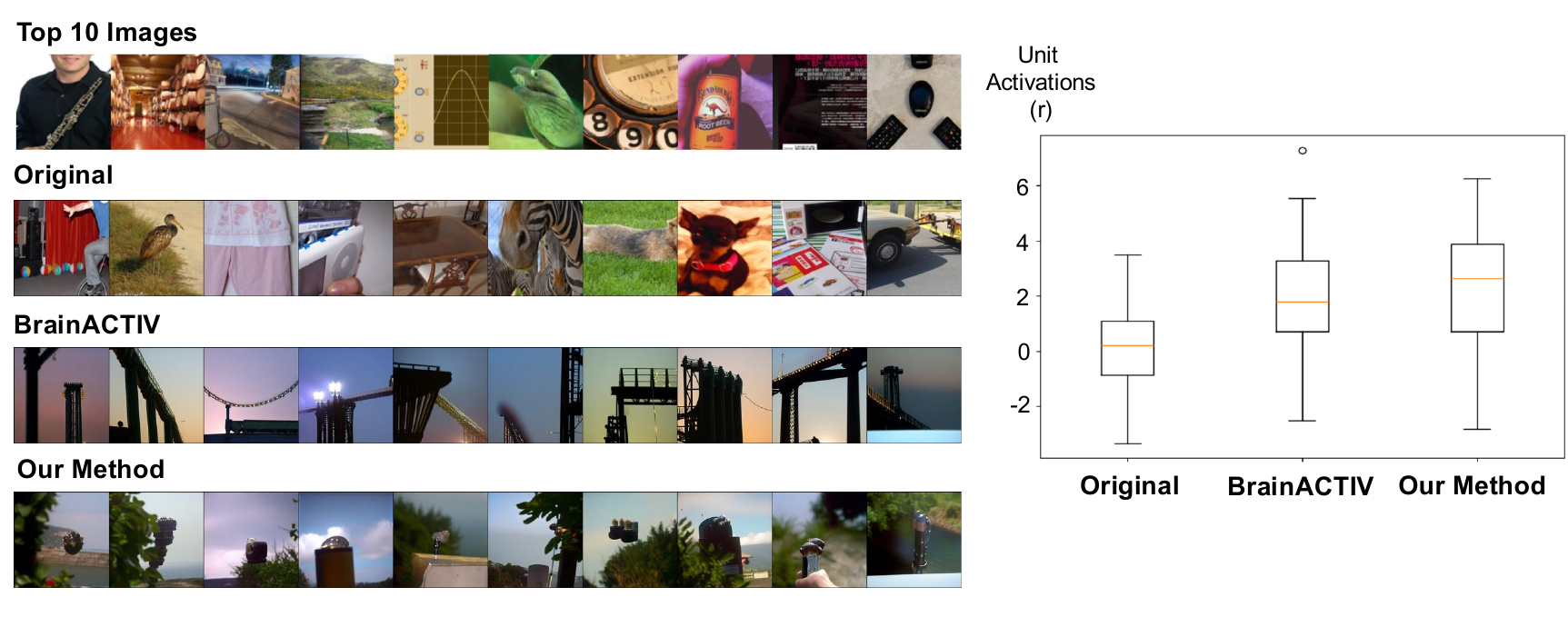}
   \caption{\textbf{One latent modification example to generate preferred images with increasing unit activations on ImageNet-mini Dataset.} Though the top 10 images can not demonstrate a notable pattern represented by the unit in the early layer of SimCLR model, our method can generate a uniform representation and compared with original images from ImageNet-mini, generated preferred images based on modifying the latent to the positive prior mean shows a more apparent change in activation compared with original images and representations from BrainACTIV.}
   \label{fig:ImageNet_7}
\end{figure}

\section{Experimental Settings}
\label{sec:Experimental Settings}
\subsection{Dataset Details}
We will show how the dataset was processed in this part.

\textbf{Natural Scenes Dataset (NSD)}: 
~\citep{Allen2022}: we selected 1000 images presented to Subject 1 from the dataset as our test set, while using CLIP embeddings of 9000 images shown to same subject for training. We used same test images selected by the BrainACTIV. And for the measurement of activation change and  FID score, we use all 1000 images presented to same subject as test set in COCO dataset~\citep{lin2015microsoftcococommonobjects} and their CLIP embeddings. 2k images for FID\citep{FID} calculation are the combination of 1k preferred images to maximize/minimize the activations, respectively. The CLIP embeddings are retrieved from the first output of CLIP encoder~\citep{radford2021clip}, the dimension is 1024 for every image.

\textbf{Imagenet-mini}~\citep{5206848}:
The train and test split follow the default settings in the orginal dataset, which divided the whole dataset into train/val sets, we use train as our train set and val as our testset. Train set consists of 34,745 images, and test set consists of 3,923 images.
For the training process of SimCLR, we used following transform techniques in torchvision.transforms to edit the orginal Imagenet-mini images:

\begin{lstlisting}[language=Python]
transforms.RandomResizedCrop(128, scale=(0.08, 1.0), 
interpolation=Image.BICUBIC)
transforms.RandomHorizontalFlip()
transforms.ToTensor()
\end{lstlisting}

\subsection{Model Implementation Details}
\subsubsection{VAE Implementation and Training}

\paragraph{VAE:} For \textbf{NSD} and \textbf{ImageNet} dataset to learn CLIP embeddings, using VAE architecture~\citep{kingma2013auto} consists of an encoder and a decoder with the following layers:

\begin{itemize}
  \item \textbf{Encoder (CLIP $\rightarrow$ latent)}
        \begin{itemize}
          \item FC\textsubscript{enc}: $1024 \longrightarrow 4096$  
                (first 2048\,dims are $\mu$, last 2048\,dims are $\log\sigma^{2}$)
        \end{itemize}

  \item \textbf{Latent layer}
        \begin{itemize}
          \item $\mu\in\mathbb{R}^{2048}$,\; $\log\sigma^{2}\in\mathbb{R}^{2048}$
          \item Reparameterisation: $z=\mu+\epsilon\odot\exp(0.5\,\log\sigma^{2})$
        \end{itemize}

  \item \textbf{Decoder (latent $\rightarrow$ CLIP)}
        \begin{itemize}
          \item FC\textsubscript{dec}: $2048 \longrightarrow 1024$  
                (reconstructed CLIP embedding)
        \end{itemize}
\end{itemize}

This CLIP-to-CLIP VAE is distinct from the image-space VAE used in the latent optimization experiments, whose decoder produces images directly.

The VAE training configurations are shown in the Table \ref{tab:VAE_train_appendix}.

\subsubsection{Implementation of Diffusion Pipeline}
All images are first converted to a 1024-dimensional representation with the frozen CLIP-ViT-H/14 encoder released by LAION~\citep{schuhmann2022laionb}.  For every split of the Natural Scenes Dataset we iterate once over the raw‐pixel loader, compute the embeddings and cache the resulting CLIP features.  Subsequent experiments therefore operate purely in embedding space and never revisit the expensive vision backbone.

Image synthesis uses the “img-to-img’’ variant of Stable Diffusion v1.5 from the \textsc{Diffusers} library~\citep{Leocadio_2025}.  We swap the original PNDM scheduler for DDIM (better speed/quality trade-off) and disable the safety checker to avoid unintended filtering.  Every network in the pipeline—U-Net, text encoder, VAE, and the IP-Adapter~\citep{ye2023ipadaptertextcompatibleimage} that injects the CLIP embedding into the cross-attention blocks—remains entirely frozen.  We use an IP-Adapter checkpoint compatible with the ViT-H/14 image embeddings employed for feature extraction.

During inference each RGB frame ($512\times512$ px) is supplied together with a target CLIP embedding—either the baseline embedding obtained by ridge regression or the one produced by our activation-aware VAE.  Unless noted otherwise we set the SDEdit strength to $\gamma=1.0$; We run 50 DDIM steps per image, draw a single sample, and fix the random seed to 42 for strict comparability across all methods.

\subsubsection{Implementation of Other Models}
\paragraph{SimCLR} 
We train a SimCLR encoder with a ResNet-50 backbone on ImageNet-mini, as described in Section~\ref{subsubsec:Activate Unit in Artificial Neural Networks}. Training uses the $128\times128$ image crops and augmentations specified above, with the model adjusted to accommodate this input resolution.

\begin{table*}[ht]
\centering
\caption{VAE Hyperparameters and Training Settings for NSD and ImageNet-mini datasets.}
\label{tab:vae_hyperparams_appendix}
\small
\begin{tabularx}{\textwidth}{@{}l*{2}{>{\raggedright\arraybackslash}X}@{}}
\toprule
\textbf{Hyperparameter}  & \textbf{NSD} & \textbf{ImageNet}  \\ 
\midrule
\multicolumn{3}{l}{\textbf{Data Hyperparameters}} \\
\midrule
\text{input\_size} & 1024 dim embeddings & 1024 dim embeddings  \\
\text{n\_channels} & 1 & 1 \\
\midrule
\multicolumn{3}{l}{\textbf{Model Hyperparameters}} \\
\midrule
\text{latent\_dim} & 2048 & 2048 \\
\midrule
\multicolumn{3}{l}{\textbf{Training Settings}} \\
\midrule
\text{seed} & 0 & 0\\
\text{batch\_size} & 128 &128 \\
\text{epochs}  & 150 & 150  \\
\text{optimizer} & Adam & Adam  \\
\text{lr}  & $1 \times 10^{-4}$ & $1 \times 10^{-3}$ \\
\text{scheduler} &  StepLR (step\_size=10, $\gamma=0.9$) &StepLR (step\_size=10, $\gamma=0.9$)  \\
\text{grad\_clip} & 1.0 &1.0 \\
\bottomrule
\label{tab:VAE_train_appendix}
\end{tabularx}
\end{table*}

\subsection{Metric Details}

\paragraph{Quantifying Similarity} To further evaluate the structured landscape, we used similarity and realism metrics. 
For similarity between original images and images with activation optimized but unrelated features preserved, we use cosine similarity as follows. Given a set of images $\{I_i\}$ with feature embeddings $\{\mathbf{f}_i\}$, the similarity score is expressed as:
\[
\text{Cosine Similarity}(\mathbf{f}_i, \mathbf{f}_j) = \frac{\mathbf{f}_i \cdot \mathbf{f}_j}{\|\mathbf{f}_i\| \|\mathbf{f}_j\|}
\]

\paragraph{Implementation of Similarity Calculation:} 
Following Section~\ref{subsubsec:Optimizing Unit Activations while Preserving Local Features}, feature preservation is quantified using a pretrained \textbf{ResNet18} encoder~\citep{he2016deep}. We compute cosine similarity between the feature embedding of each original image and that of its corresponding optimized image. Higher similarity indicates greater preservation of image features.

\paragraph{Quantifying Realism: Frechét Inception Distance (FID):}
To quantify overall realism we report the Frechét Inception Distance, computed on the 2 048-D pool-3 activations of a frozen Inception-v3 network.
Let $\mu_r$, $\Sigma_r$ and $\mu_g$, $\Sigma_g$ be the empirical means and covariances of the real and generated image features, respectively.  The FID is

\[
\operatorname{FID}
= \lVert \mu_r - \mu_g \rVert_2^{\,2}
+ \operatorname{Tr}\!\bigl(\Sigma_r + \Sigma_g
- 2\,(\Sigma_r \Sigma_g)^{1/2}\bigr).
\]
Lower values indicate that the generated distribution is closer to the real one.
For the NSD evaluation, each method generates 1,000 images in the positive direction and 1,000 in the negative direction. We pool these into a single set of 2,000 generated images per method and compute one FID against the 1,000 original test images repeated twice. FID is computed with the \textsc{torchmetrics} implementation using \texttt{feature=2048}.

\subsection{Computational Resources}

All experiments were carried out on a single workstation equipped with
\textbf{8 × NVIDIA RTX A5000 GPUs} (each with 24 GB on-board memory, CUDA 12.2, driver 535.183),
an AMD EPYC 7543 CPU (32 cores, 2.8 GHz) and 512 GB of system RAM.
One GPU was used for model training or inference at a time.
Training a CLIP-to-CLIP VAE on the NSD train split ($\approx$ 9 k images) with a batch-size of 128
took $\approx$ 6.5 hours on a single A5000; Optimizing on Imagenet-mini completed in under
40 minutes.
For diffusion-based generation (IP-Adapter + Stable‐Diffusion v1.5, 50 DDIM steps) ,producing one full 2 k-image
evaluation set (positives + negatives) took about 70 minutes.

\clearpage

\section{Implementation Details}
\label{sec:Implementation Details}
\subsection{Training Details}
The following pseudocode summarizes the training objective and generation procedure described in Section~\ref{sec:training_and_generation}.

\begin{algorithm}[H]
\caption{VAE Training with Activation Regularization}
\label{alg:vae_training_activation_appendix}
\begin{algorithmic}[1]
\Require
\State \texttt{vae}: Variational Autoencoder model
\State \texttt{optimizer}: Optimizer (Adam)
\State \texttt{scheduler}: Learning rate scheduler (StepLR)
\State \texttt{num\_epochs}: Number of training epochs
\Ensure
\State VAE with structured latent space and high-quality reconstruction 
\Statex
\Procedure{TrainVAE}{\texttt{vae}, \texttt{optimizer}, \ldots}
    \For{epoch = 1 to \texttt{num\_epochs}}
        \For{each batch}
            \State \texttt{optimizer.zero\_grad()}
            \State \textbf{Forward Pass:}
            \State $(\mu,\log\sigma^2) \gets \texttt{vae.encode}(\texttt{data})$
            \State $\eta \sim \mathcal{N}(0,I)$;\quad $z \gets \mu+\exp(\tfrac{1}{2}\log\sigma^2)\odot\eta$
            \State $\hat{x} \gets \texttt{vae.decode}(z)$
            \State \textbf{Compute Loss:}
            \State $\text{Reconstruction Loss} \gets \lVert \texttt{data}-\hat{x}\rVert_2^2$
            \State \textbf{Compute Prior Parameters:}
            \State $\sigma^{\text{prior}} \gets \left| \dfrac{\lambda}{\texttt{activation\_values} + \epsilon} \right|$
      \State $\mu^{\text{prior}}_i \gets
             \begin{cases}
               \mu_{\text{pos}}, & \text{if } \texttt{activation\_values}_i>0,\\
               \mu_{\text{neg}}, & \text{otherwise}.
             \end{cases}$
            \State Compute $\text{KLD Loss}$ from $\mu,\sigma,\mu^{\text{prior}},\sigma^{\text{prior}}$ using Eq.~\ref{eq:kld_modified}
            \State $\text{Total Loss} \gets \omega_{\text{recon}} \times \text{Reconstruction Loss} + \omega_{\text{KLD}} \times \text{KLD Loss}$
            \State \textbf{Backpropagation:}
            \State $\text{Total Loss.backward()}$
            \State \texttt{optimizer.step()}
        \EndFor
        \State \texttt{scheduler.step()}
    \EndFor
\EndProcedure
\end{algorithmic}
\end{algorithm}

\paragraph{Latent Modification and Image Generation}

To generate high-quality preferred images that enhance activations based on the latent modification, the algorithm is shown as following block~\ref{alg:latent_modification}:
\begin{algorithm}[H]
\caption{\modelname{}: Preferred–Image Generation}
\label{alg:latent_modification}
\begin{algorithmic}[1]
\Require
    \Statex $i_{\text{in}}$ … RGB input image  
    \Statex $\texttt{clip\_enc}$ … frozen CLIP–ViT-H encoder  
    \Statex $x$ … 1024-d CLIP embedding of $i_{\text{in}}$  
    \Statex $\texttt{vae}$ … activation–aware CLIP$\!\rightarrow$CLIP VAE  
    \Statex $z,\mu$ … latent code / mean; $\mu_{\text{pos}},\mu_{\text{neg}}$ … activation priors  
    \Statex $\texttt{ip\_adapter}$ … frozen IP-Adapter  
    \Statex $\texttt{sd15}$ … Stable-Diffusion v1.5 (DDIM)  
    \Statex $\alpha$ … linear interpolation weight, $\gamma$ … SDEdit strength  
    \Statex $r_{\text{dir}}\!\in\!\{\mathrm{pos},\mathrm{neg}\}$ … desired ROI shift
\Ensure preferred image $i_{\text{out}}$
\Procedure{GenPreferred}{$i_{\text{in}}, r_{\text{dir}}$}
    \State $x \gets \texttt{clip\_enc}(i_{\text{in}})$
    \State $(\mu,\log\sigma^{2}) \gets \texttt{vae.encode}(x)$;\; $z\!\gets\!\mu$
    \If{$r_{\text{dir}}=\mathrm{pos}$}
        \State $\mu_{\text{prior}} \gets \mu_{\text{pos}}$
    \Else
        \State $\mu_{\text{prior}} \gets \mu_{\text{neg}}$
    \EndIf
    \State $z' \gets \alpha\mu_{\text{prior}}+(1-\alpha)z$
    \State $x_{\text{gen}} \gets \texttt{vae.decode}(z')$
    \State $i_{\text{out}} \gets \texttt{ip\_adapter.generate}
           (\text{clip}=x_{\text{gen}},\,\text{image}=i_{\text{in}},\,
             \gamma,\,50\text{ steps})$
    \State \Return $i_{\text{out}}$
\EndProcedure
\end{algorithmic}
\end{algorithm}
Here are more detailed explanations of our generation process:
\begin{itemize}
\item The input image $i_{\text{in}}$ is projected to a CLIP embedding x with the frozen encoder.
\item The activation-aware VAE encodes x to latent mean $\mu$; we take $z=\mu$ for a deterministic edit.
\item Following Section~\ref{subsubsec:Manipulate Activations in Human Brain}, the latent edit is $z'=\alpha\mu_{\text{prior}}+(1-\alpha)z$, where $\mu_{\text{prior}}$ is the positive prior $\mu_{\text{pos}}$ to raise activity or the negative prior $\mu_{\text{neg}}$ to suppress it.
\item Decoding z’ gives an intermediate CLIP embedding $x_{\text{gen}}$.
\item $x_{\text{gen}}$ is injected—via a frozen IP-Adapter—into Stable-Diffusion v1.5.
With SDEdit strength $\gamma=1.0$ the reverse DDIM process relies entirely on $x_{\text{gen}}$; smaller $\gamma$ blends in low-level structure from the reference image.
\item The sampler returns the final preferred image $i_{\text{out}}$, which is used in all downstream analyses.
\end{itemize}
\subsection{Details about Optimizing Unit Activations with Latent}

The algorithm for optimizing neural network unit responses using latent variables from a VAE is listed as following algorithm block~\ref{alg:optimize_activation_appendix}:

\begin{algorithm}[H]
\small
\caption{Optimize Activation with Latent Variables}
\label{alg:optimize_activation_appendix}
\begin{algorithmic}[1]
\interlinepenalty1000
\Require
\State \texttt{net}: Pre-trained neural network model
\State \texttt{vae}: Pre-trained Variational Autoencoder
\State \texttt{layer\_name}: Target layer in \texttt{net}
\State \texttt{unit\_idx}: Index of target unit
\State \texttt{num\_iterations}: Number of optimization steps
\State \texttt{learning\_rate}: Learning rate for optimizer
\State \texttt{KLD\_weight}: Weight for Kullback-Leibler Divergence loss
\State \texttt{target}: Direction ``max" or ``min")
\State \texttt{init\_data}: Initial data sample
\State $\mathcal{J}$: Latent dimensions associated with the target unit
\Ensure
\State Optimized responses scores, Controlled KLD loss
\Statex
\Procedure{OptimizeActivation}{$\texttt{net}, \texttt{vae}, \ldots$}
    \State \textbf{Initialize Latent Variable}:
        \State $(\mu, \logvar) \gets \texttt{vae.encode}(\texttt{init\_data})$
        \State $z \gets \mu$;\quad $z_0 \gets \mu$

    \State \textbf{Set Up Optimizer}:
    \State \texttt{optimizer} $\gets$ Adam($z$, lr=$\texttt{learning\_rate}$)
    
    \State \textbf{Register Activation Hook} on \texttt{layer\_name}
    
    \For{iteration = 1 to \texttt{num\_iterations}}
        \State \textbf{Decode Latent to Image}:
        \State $x \gets \texttt{vae.decode}(z)$
        
        \State \textbf{Forward Pass}:
        \State \texttt{net}(x)
        
        \State \textbf{Compute Loss}:
        \If{$\texttt{target} = \text{``max"}$}
            \State $\text{loss} \gets -\text{activation} + \omega_{\text{KLD}} \times \text{KLD}(z)$
        \Else
            \State $\text{loss} \gets \text{activation} + \omega_{\text{KLD}} \times \text{KLD}(z)$
        \EndIf
        
        \State \textbf{Backpropagation}:
        \State \text{optimizer.zero\_grad}()
        \State $\text{loss.backward()}$
        \State \text{optimizer.step}()
        \State $z_j \gets (z_0)_j$ for all $j\notin\mathcal{J}$
    \EndFor
    
    \State \textbf{Remove Activation Hook}
    \State \textbf{Return} $x$, activation scores
\EndProcedure
\end{algorithmic}
\end{algorithm}
\paragraph{More Explanation of the Algorithm}

This algorithm describes the image-space VAE optimization experiment in Section~\ref{subsubsec:Optimizing Unit Activations while Preserving Local Features}. To preserve unrelated features, only the latent dimensions $\mathcal{J}$ associated with the selected unit are updated; the remaining coordinates stay at their initial values. The VAE and target network remain fixed while the latent code is optimized to increase or decrease the target response.

\textbf{Initialization:} We initialize the latent variable \( z \) by encoding the input image using the VAE encoder:

\[
(\mu, \logvar) \gets \texttt{vae.encode}(\texttt{init\_data})
\]
\[
z \gets \mu
\]

\textbf{Optimization Loop:} For each iteration, we perform the following steps:

1. \textbf{Decode Latent to Image:} The current latent variable \( z \) is decoded to generate an image \( x \):

   \[
   x \gets \texttt{vae.decode}(z)
   \]

2. \textbf{Forward Pass and Activation Capture:} The generated image \( x \) is passed through the neural network \( \texttt{net} \). We utilize a registered \textbf{hook function} on the target layer to capture the activation of the target unit during this forward pass:

   \[
   \texttt{net}(x)
   \]

   The hook function retrieves the target activation \( \text{activation} \), which will be used in the loss computation. This approach allows us to monitor and manipulate the activation of specific units without altering the forward pass logic of the network.

3. \textbf{Compute Loss:} We compute the loss function, which consists of two components: the activation term and the Kullback-Leibler Divergence (KLD) regularization term.

   - \textbf{Activation Term:} Depending on the optimization direction (maximize or minimize), we use the negative or positive of the target unit's activation:

     \[
     \text{loss} = 
     \begin{cases}
     -\text{activation}, & \text{if } \texttt{target} = \text{``max"} \\
     \phantom{-}\text{activation}, & \text{if } \texttt{target} = \text{``min"}
     \end{cases}
     \]

   - \textbf{KLD Regularization Term:} The KLD term ensures that the latent variable \( z \) remains within a plausible region of the latent space.

     We consider two method for the KLD calculation. One is \textbf{our method}, and the other one is using the latent space of \textbf{Standard VAE} to compute KLD loss, which is our \textbf{baseline}.

     \begin{itemize}
         \item \textbf{Our Method:} We calculate the KLD using a modified prior that incorporates the desired change in activation. Specifically, the prior mean \( \mu^{\text{prior}} \) is adjusted based on the target direction:

           \[
           \mu^{\text{prior}} = 
           \begin{cases}
           +1, & \text{if } \texttt{target} = \text{``max"} \\
           -1, & \text{if } \texttt{target} = \text{``min"}
           \end{cases}
           \]

           The KLD is then computed as:

            \[
            \begin{aligned}
            \text{KLD}_{\text{our}}(z) &= \frac{1}{2} \sum_{i=1}^{d_z} \bigg[ 2\log \left( \frac{\sigma_i^{\text{prior}}}{\sigma_i} \right) - 1 \\
            & \quad + \frac{\sigma_i^2 + (\mu_i - \mu_i^{\text{prior}})^2}{(\sigma_i^{\text{prior}})^2} \bigg]
            \end{aligned}
            \]
           where \( \sigma_i^{\text{prior}} \) is set to a small constant (e.g., \( 0.1 \)).

         \item \textbf{Standard VAE:} We calculate the KLD using the standard Gaussian prior with zero mean and unit variance:

           \[
           \text{KLD}_{\text{standard}}(z) = \frac{1}{2} \sum_{i=1}^{d_z} \left( \mu_i^2 + \sigma_i^2 - \log(\sigma_i^2) - 1 \right)
           \]

     \end{itemize}

   - \textbf{Total Objective Function:} The total loss combines the activation loss and the KLD regularization:

     \[
     \text{loss} = \text{activation\_loss} + \omega_{\text{KLD}} \times \text{KLD}(z)
     \]

     where \( \text{KLD}(z) \) is either \( \text{KLD}_{\text{our}}(z) \) or \( \text{KLD}_{\text{standard}}(z) \) depending on the mode.

4. \textbf{Backpropagation:} We perform backpropagation to compute the gradients of the loss with respect to \( z \) and update \( z \) using the optimizer.

   \[
   \text{optimizer.zero\_grad}()
   \]
   \[
   \text{loss.backward()}
   \]
   \[
   \text{optimizer.step}()
   \]

\textbf{Configurations:} During the implementation, learning rate was set to 0.1, num\_iterations to 10000 and $\omega_{\text{KLD}}$ to 0.001.

\section{Predictive Power of In-silico brain simulator}
\label{sec:Predictive Power}
The following plots compare observed and predicted responses for the in-silico brain simulator described in Section~\ref{subsec:Models}. The simulator uses a frozen DINO-ViT feature extractor and a single-layer ViT ensemble over its 12 layer outputs. Separate simulators are trained for different ROI groups. In these fMRI plots, the displayed correlation is the arithmetic mean of the Pearson correlations computed separately for the ROIs in each group.

\begin{figure}[H]
    \centering
    \includegraphics[width=\linewidth,height=0.80\textheight,keepaspectratio]{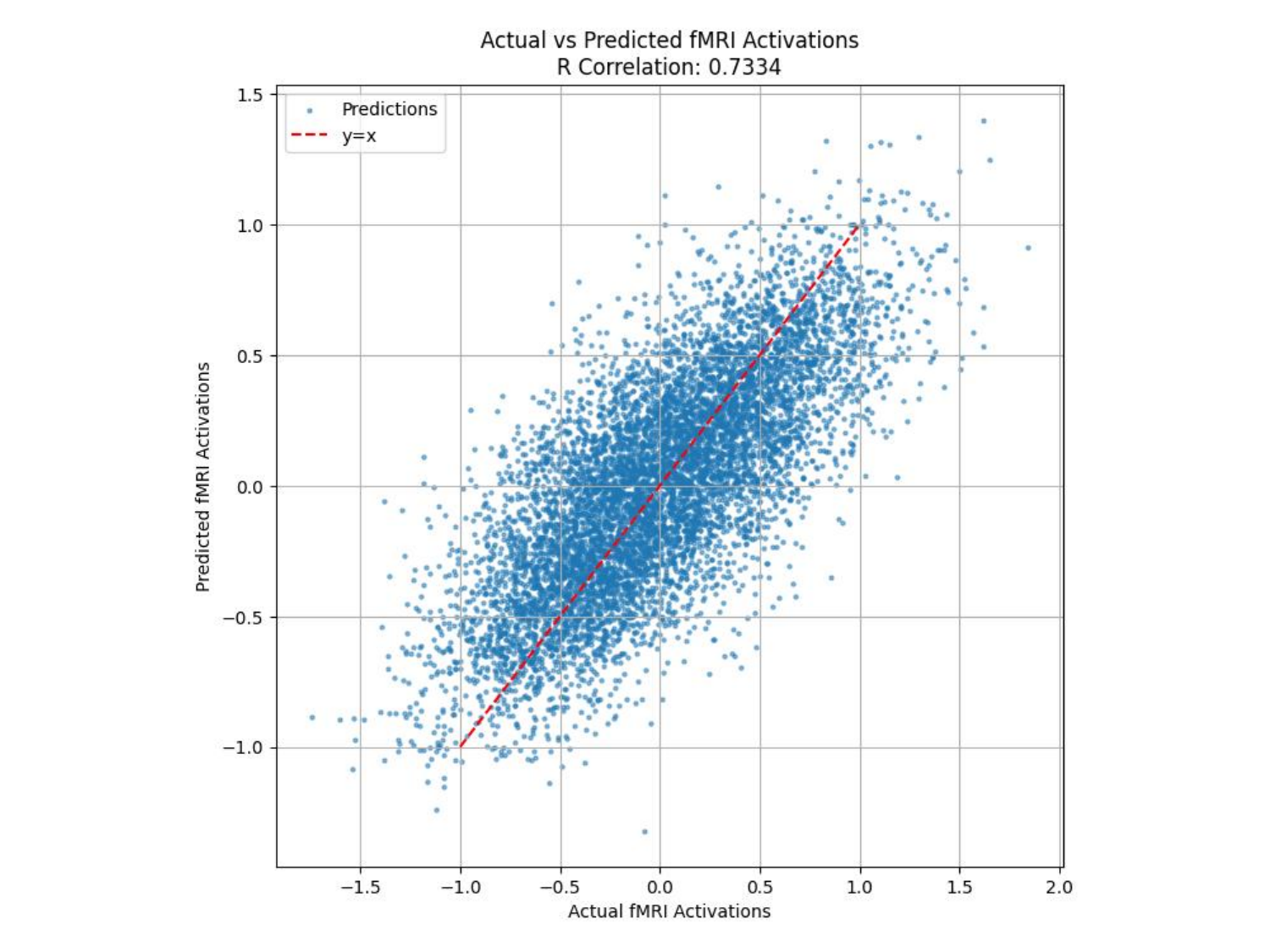}
    \caption{Observed versus predicted responses. ROIs: OFA, FFA-1, EBA, VWFA-1, OPA, PPA, RSC, OWFA.}
\end{figure}
\begin{figure}[H]
    \centering
    \includegraphics[width=\linewidth,height=0.80\textheight,keepaspectratio]{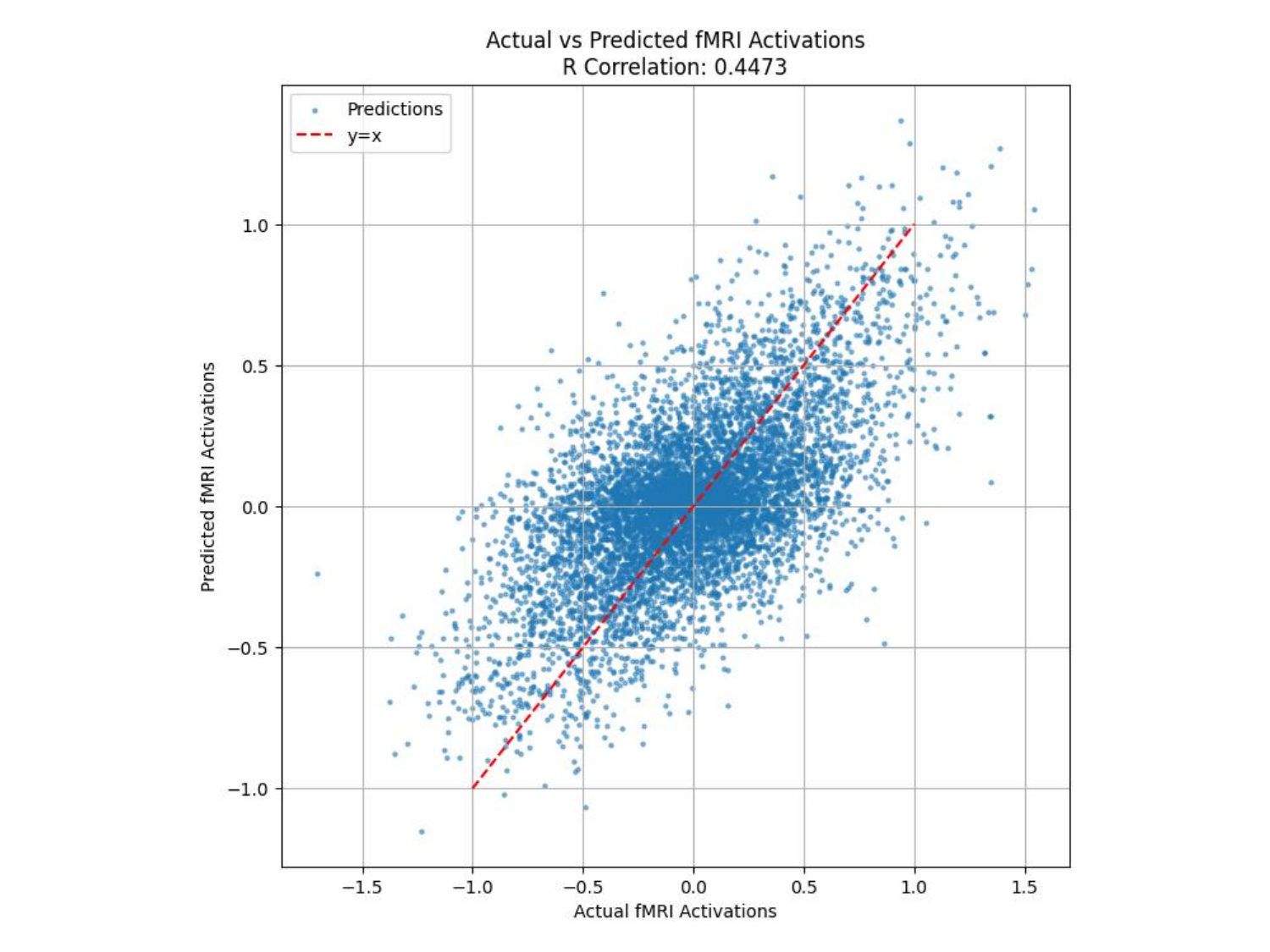}
    \caption{Observed versus predicted responses. ROIs: IPS4, SPL1, TO1, PHC2, VO2, PGi, VVC, TE2a.}
\end{figure}

\begin{figure}[H]
    \centering
    \includegraphics[width=\linewidth,height=0.80\textheight,keepaspectratio]{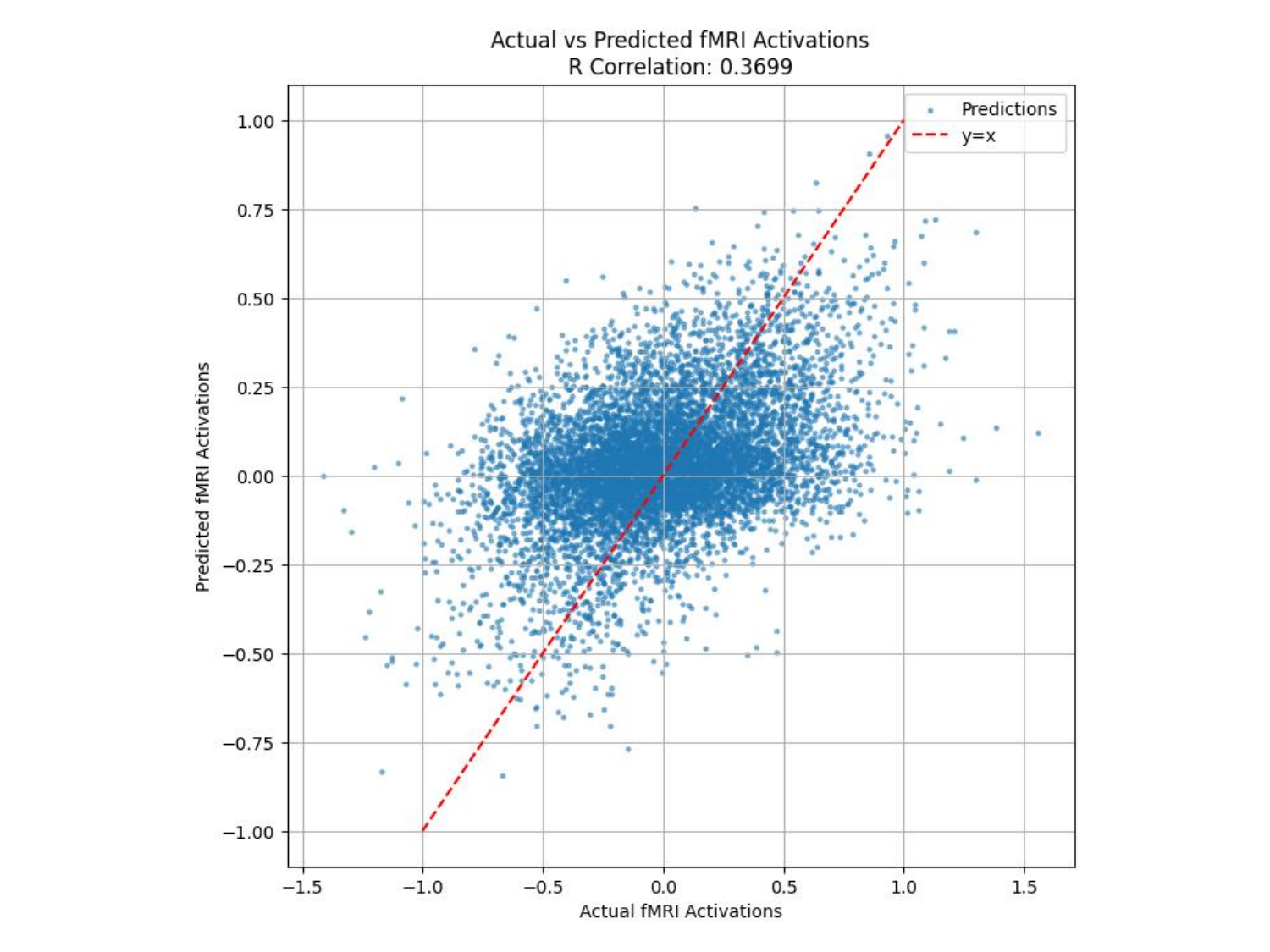}
    \caption{Observed versus predicted responses. ROIs: STSdp, 31a, 31pd, DVT, FST, Gp, PGs, PH.}
\end{figure}

\begin{figure}[H]
    \centering
    \includegraphics[width=\linewidth,height=0.80\textheight,keepaspectratio]{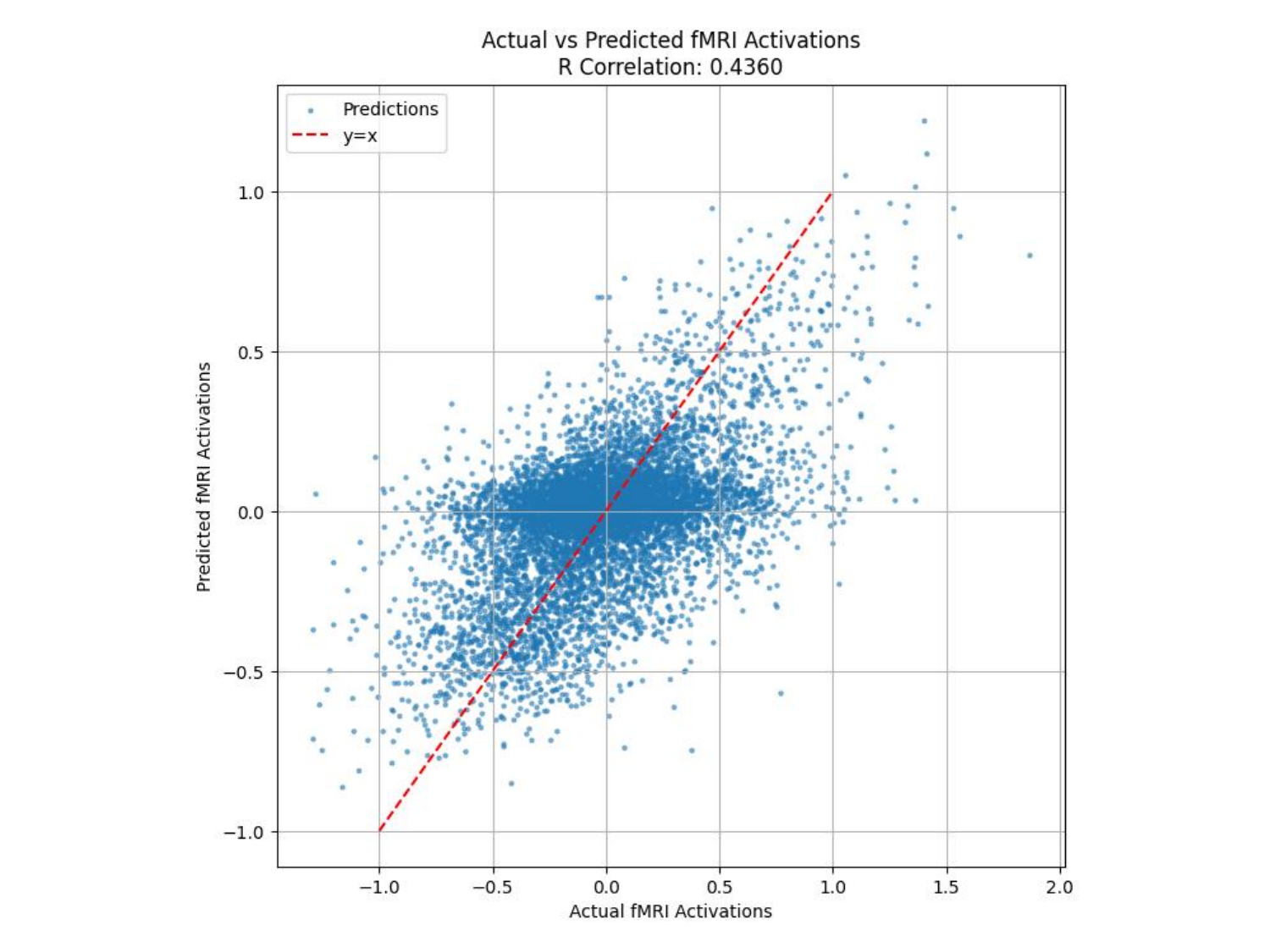}
    \caption{Observed versus predicted responses. ROIs: TE2p, PHT, TPOJ1, TPOJ2, TPOJ3, IP0, IP1, IP2.}
\end{figure}

\begin{figure}[H]
    \centering
    \includegraphics[width=\linewidth,height=0.80\textheight,keepaspectratio]{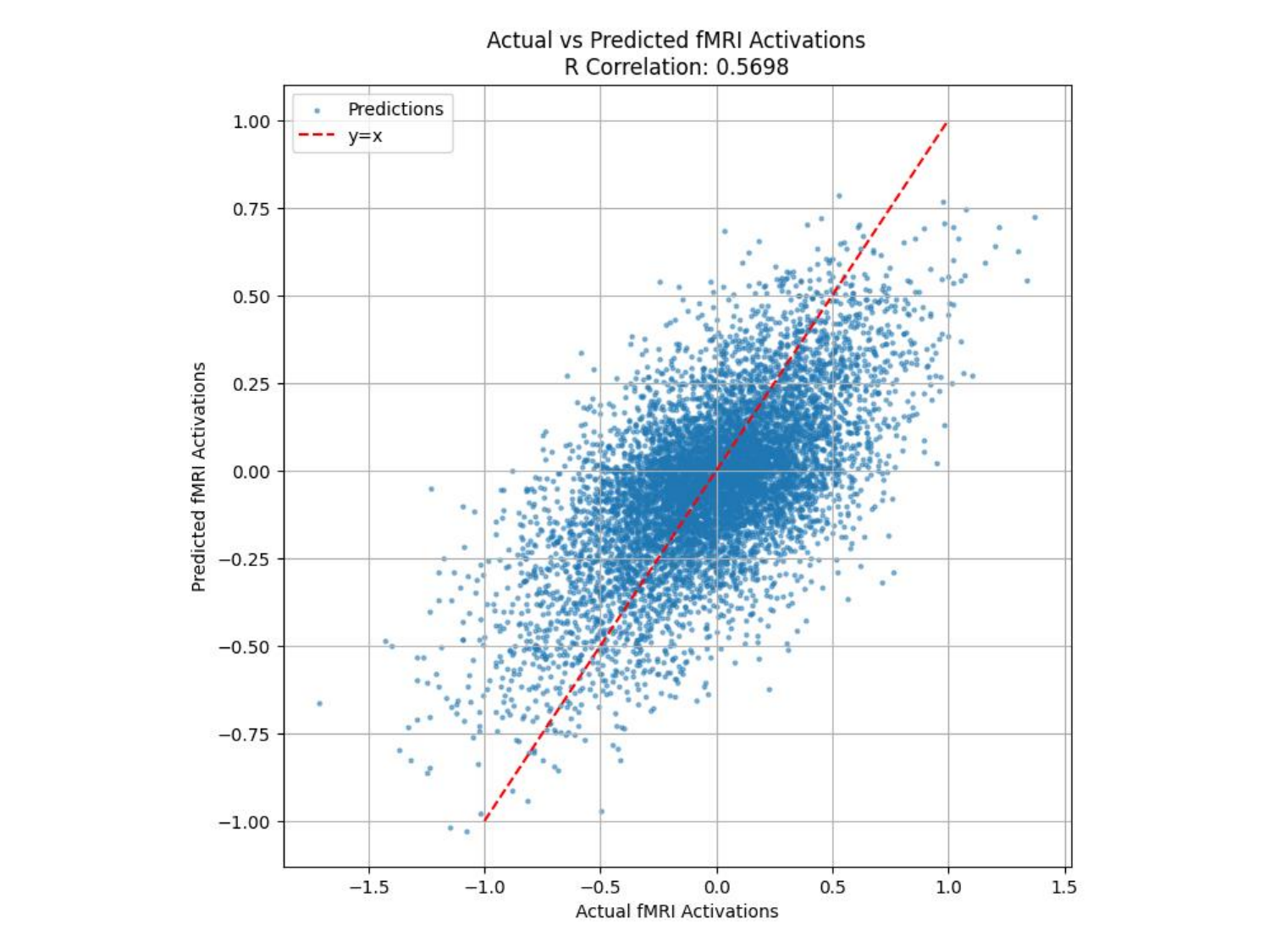}
    \caption{Observed versus predicted responses. ROIs: V1, V2, V3, V4, V8, LO1, LO2, V4t.}
\end{figure}

\begin{figure}[H]
    \centering
    \includegraphics[width=\linewidth,height=0.80\textheight,keepaspectratio]{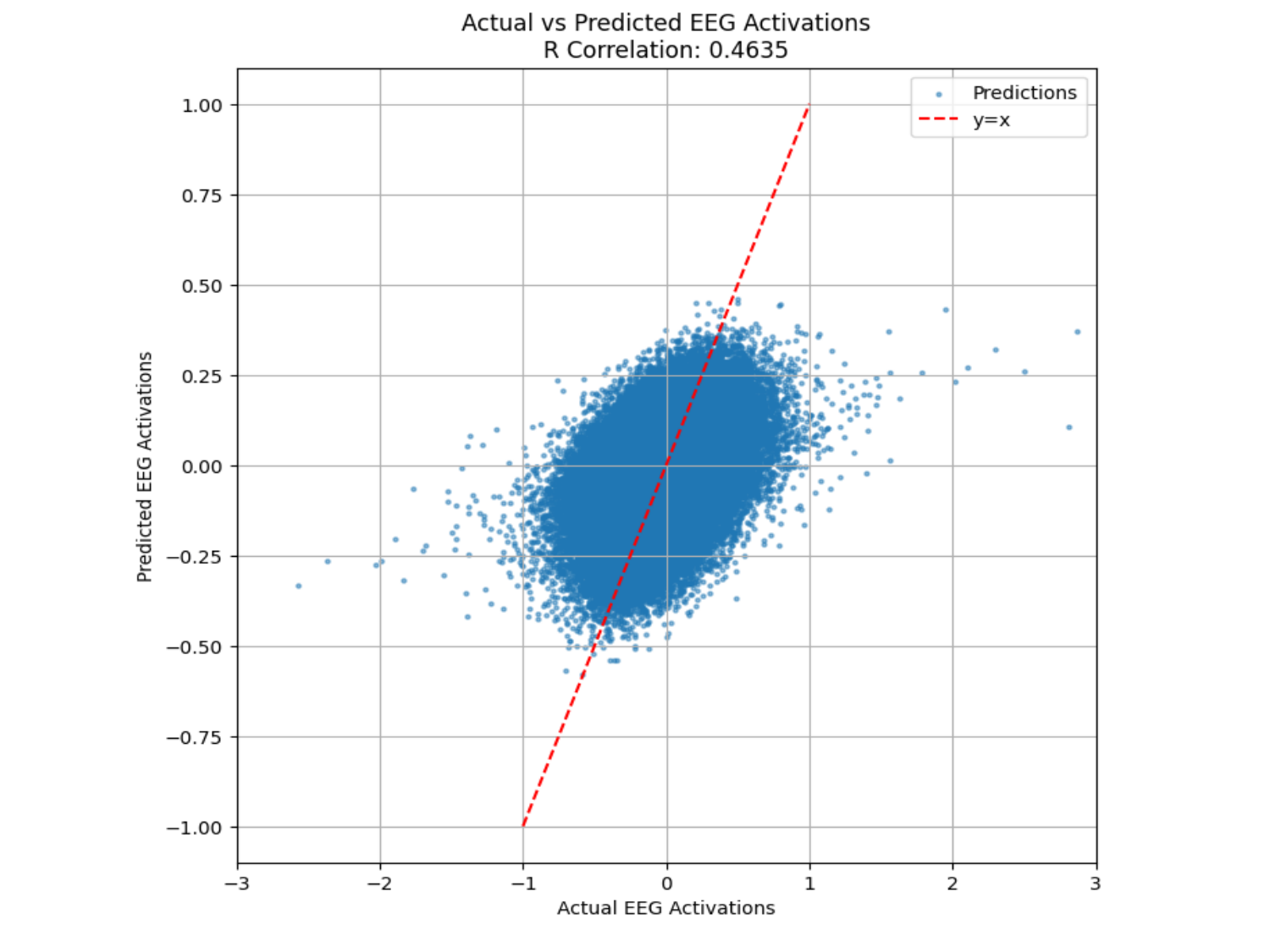}
    \caption{Observed versus predicted responses. Occipital Channels: O1, Oz, O2, PO7, PO3, POz, PO4, PO8.}
\end{figure}

\clearpage
\end{document}